\documentclass[leqno,11pt]{article}
\usepackage{CJKutf8}
\usepackage[utf8]{inputenc}
\usepackage[OT2, T1]{fontenc}
\usepackage[russian, english]{babel}
\usepackage{glottometrics}
\usepackage{float}
\usepackage{multirow}
\usepackage{subcaption}

\usepackage{csquotes}
\DeclareMathOperator{\expect}{\mathbb{E}}

\usepackage{amsthm}

\newtheorem{property}{Property}[section]

\newcommand{\repository}[0]{https://github.com/IQL-course/IQL-Research-Project-21-22}

\usepackage[toc,page,title]{appendix} 
\usepackage{etoolbox}

\newtoggle{anonymous}
\togglefalse{anonymous}

\glottovol{XX}
\glottoyear{20XX}
\glottodoi{Here will be added DOI number by Glottometrics.}

\iftoggle{anonymous}{
\runningauthors{}
}
{
\runningauthors{Petrini, Casas-i-Muñoz, Cluet-i-Martinell, Wang, Bentz \& Ferrer-i-Cancho}
}

\runningtitle{Direct and indirect evidence of compression of word lengths.}

\title{Direct and indirect evidence of compression of word lengths. \\ 
Zipf's law of abbreviation revisited.}

\iftoggle{anonymous}{
\addauthor{1}{Some author}{Some ORCID}
\affil{1}{Some affiliation.}

}{
\addauthor{1}{Sonia Petrini}{0000-0002-0514-6223}
\addauthor{2}{Antoni Casas-i-Muñoz}{0000-0001-5690-316X}
\addauthor{2}{Jordi Cluet-i-Martinell}{0000-0003-4188-6728}
\addauthor{2}{Mengxue Wang}{0000-0002-8262-9333}
\addauthor{3}{Christian Bentz}{0000-0001-6570-9326}
\addauthor{1*}{Ramon Ferrer-i-Cancho}{0000-0002-7820-923X}

\affil{1}{
Quantitative, Mathematical and Computational Linguistics Research Group. Departament de Ci\`encies de la Computaci\'o, Universitat Polit\`ecnica de Catalunya (UPC), Barcelona, Catalonia, Spain.}
\affil{2}{Universitat Polit\`ecnica de Catalunya (UPC), Barcelona School of Informatics, Barcelona, Catalonia, Spain.}
\affil{3}{Department of Linguistics, University of Tübingen, Tübingen, Germany.}

\correspond{*}{Corresponding author's email: rferrericancho@cs.upc.edu}

}

\begin{document}
\maketitle

\renewcommand{\sectionautorefname}{Section}
\renewcommand{\subsectionautorefname}{Section}
\renewcommand{\subsubsectionautorefname}{Section}


\begin{abstract}
Zipf's law of abbreviation, the tendency of more frequent words to be shorter, is one of the most solid candidates for a linguistic universal, in the sense that it has the potential for being exceptionless or with a number of exceptions that is vanishingly small compared to the number of languages on Earth.  
Since Zipf's pioneering research, this law has been viewed as a manifestation of a universal principle of communication, i.e. the minimization of word lengths, to reduce the effort of communication. 
Here we revisit the concordance of written language with the law of abbreviation. Crucially, we provide wider evidence that the law holds also in speech (when word length is
measured in time), in particular in 46 languages from 14 linguistic families. Agreement with the law of abbreviation provides indirect evidence of compression of languages via the theoretical argument that the law of abbreviation is a prediction of optimal coding. Motivated by the need of direct evidence of compression, we derive a simple formula for a random baseline
indicating that word lengths are systematically below chance, across linguistic families and writing systems, and independently of the unit of measurement (length in characters or duration in time). 
Our work paves the way to measure and compare the degree of optimality of word lengths in  languages. 
\end{abstract}




\section{Introduction}
\label{sec:introduction}

It has been argued that linguistic universals are a myth \parencite{Evans2009a}, but this neglects the statistical regularities that the quantitative linguistic community has been investigating for many decades. A salient case is Zipf's law of abbreviation, the tendency of more frequent words to be shorter \parencite{Zipf1949a}. 
It holds across language families \parencite{Piantadosi2011a,Bentz2016a,Levshina2022a,Meylan2021a, Koplenig2022a}, writing systems \parencite{Wang2015a,Sanada2008a} and modalities \parencite{Boerstell2016a,Torre2019a,Hernandez2022a}, and also when word length in characters is replaced by word duration in time \parencite{Hernandez2019a}. 
Furthermore, the number of species where a parallel of this law has been confirmed in animal communication is growing over time \parencite{Semple2021a}.\footnote{The interested reader can check the latest discoveries on this law in ``Bibliography on laws of language outside human language''  at \url{https://cqllab.upc.edu/biblio/laws/}.} 
In language sciences, research on the law of abbreviation in languages measures word length in discrete units (e.g., characters) whereas, in biology, research on the law in other species typically uses duration in time. Here, we aim to reduce the gulf that separates these two traditions by promoting research on the law of abbreviation on word durations.

G. K. Zipf believed that the law of abbreviation constituted {\em indirect} evidence of the minimization of the cost of using words \parencite{Zipf1949a}. At present, Zipf's view is supported by standard information theory and its extensions: the main argument is that the minimization of $L$, the mean word length, that is indeed a simplification of Zipf's cost function,\footnote{He  referred to the cost function as ``minimum equation'' \parencite{Zipf1949a}.} leads to the law of abbreviation \parencite{Ferrer2012d,Ferrer2019c}. Using the terminology of information theory, the minimization of mean word length is known as compression. Using the terminology of quantitative linguistics, $L$ 
is the average length of tokens from a repertoire of $n$ types, that is defined as 
\begin{equation}
L = \sum_{i=1}^n p_i l_i,
\label{eq:mean_type_length}    
\end{equation}
where $p_i$ and $l_i$ are, respectively, the probability and the length of the $i$-th type.
In practical applications, $L$ is calculated replacing $p_i$ by the relative frequency of a type, that is
$$p_i = f_i/ T,$$ 
where $f_i$ is the absolute frequency of a type and $T$ is the total number of tokens, i.e.
$$T = \sum_{i=1}^n f_i.$$
This leads to a definition of $L$ that is 
$$L = \frac{1}{T} \sum_{i=1}^n f_i l_i.$$

At present, the mathematical link between the law of abbreviation and compression has been established under the assumption that words are coded optimaly so as to minimize $L$. If words are coded optimaly, the correlation between the frequency of a word and its duration cannot be positive \parencite{Ferrer2019c}. Thus, a lack of correlation between the frequency of a word and its duration does not imply absence of compression. Furthermore, it is not a warranted assumption that languages code words optimaly. Therefore, an approach to find {\em direct} evidence of compression getting rid of the assumption of optimal coding is required.

As a first approach, one could compare the value of $L$ of a language against $L_{max}$, the maximum value that $L$ could achieve in this language. The larger the gap between $L$ and $L_{max}$, the higher the level of compression in the language. However, the problem is that $L_{max}$ can be infinite {\em a priori}. To fix that problem, one could restrict $L_{max}$ to be finite but then this raises the question of what should be the finite value of $L_{max}$ and why. For these reasons, here we resort to the notion of random baseline,
that here is defined assuming some random mapping of word types into strings. 
In previous research, the random baseline was defined by the average word length resulting from a shuffling of the current length/duration of types so as to check if $L$ was smaller than expected by chance in that random mapping \parencite{Ferrer2012d,Heesen2019a}. 
Critically, an exact method to compute the random baseline, namely the expected word length in these shufflings, is missing.

The remainder of the article is organized as follows. In \autoref{sec:baselines}, we introduce the definition of $L_r$, the random baseline, that we will use to explore direct evidence of compression. In particular, we derive a simple formula for $L_r$ that will simplify future research on compression in natural communication systems.
In \autoref{sec:material} and \autoref{sec:methodology}, we present, respectively, the materials and methods that will be used to provide further evidence of compression and the law of abbreviation in real languages with emphasis on word durations. In \autoref{sec:methodology}, we present a new unsupervised method to exclude words with foreign characters in line with good practices for research on linguistic laws and communicative efficiency \parencite{Meylan2021a}.
In \autoref{sec:results}, we show that the law of abbreviation holds without exceptions in a wide sample of languages, independently of the unit of measurement of word length, namely characters or duration in time, 
providing further indirect evidence of compression in languages.
In addition, the random baseline indicates that word lengths are systematically below chance, across linguistic families and writing systems, independently of the unit of measurement (length in characters or duration in time), providing direct evidence of compression.  
Finally, in \autoref{sec:discussion}, we discuss the findings in relation to the potential universality of the law of abbreviation and the universality of compression in languages. We also make proposals for future research.

\section{A random baseline revisited}
\label{sec:baselines}

In our statistical setting, the null hypothesis states that compression (minimization of word lengths) has no effect on word lengths. The alternative hypothesis states that compression has an effect on word lengths as Zipf hypothesized. If the null hypothesis is rejected then word lengths are shorter than expected by chance. 

\begin{table}[ht]
  \caption{\label{tab:example} Matrix indicating the frequency and length of three types. The mean type length is $L = \frac{235}{125}=1.88$. } 
  
  \begin{center} 
  \begin{tabular}{lll}
      &       &       \\ 
  $i$ & $f_i$ & $l_i$ \\ 
  \hline 
  1   &  100  & 2  \\         
  2   &  20   & 1  \\
  3   &  5    & 3
  \end{tabular}
  \end{center}
\end{table}

Consider a matrix with two columns, $f_i$ and $l_i$, that are used to compute the average word length $L$. The matrix in \autoref{tab:example} gives $L = \frac{235}{125} = 1.88$. 
We consider the null hypothesis of a random mapping of probabilities into lengths, namely that the ordering of the $f_i$'s or the $l_i$'s in \autoref{tab:example} is arbitrary and results from a random shuffling of one of these variables or both. We use $f_i'$, $l_i'$ and
$p_i'$ for the new values of $f_i$, $l_i$ and $p_i$ that result from one of these shufflings. 

This null hypothesis was introduced in research on compression in human language and animal communication to test if $L$ is significantly small using a permutation test \parencite{Ferrer2012d,Heesen2019a}. Later, it was used to estimate the degree of optimality of word lengths \parencite{Moreno2021a,Pimentel2021a}. Our new contribution here is a precise mathematical characterization of the null hypothesis and the derivation of a simple formula the expected word length. 

In the context of computing average word length, the matrix in \autoref{tab:example} is equivalent to a matrix where the column $f_i$ is replaced by a column with $p_i$ thanks to 
$$p_i' = \frac{f_i'}{T}.$$ 
Indeed, the null hypothesis has three variants 
\begin{enumerate}
    \item 
    Single column shuffling. Only the column of $f_i$ or $p_i$ is shuffled. 
    \item
    Single column shuffling. Only the column of $l_i$ is shuffled. 
    \item
    Dual column shuffling. The column of $f_i$ or $p_i$ and the column of $l_i$ are both shuffled. 
\end{enumerate}
In each of the variants, all random shufflings of a specific column are equally likely. In case of dual shuffling, the shuffling of one column is independent of the shuffling of the other column. 
The outcome of a dual shuffling on \autoref{tab:example} is shown in \autoref{tab:example_shuffling}.

\begin{table}[ht]
  \caption{\label{tab:example_shuffling} Matrix indicating the frequency and length of three types. The mean type length is $L = \frac{345}{125}=2.76$. } 
  
  \begin{center} 
  \begin{tabular}{lll}
      &       &       \\ 
  $i$ & $f_i'$ & $l_i'$ \\ 
  \hline 
  1   &  20    & 2  \\         
  2   &  100   & 3  \\
  3   &  5     & 1
  \end{tabular}
  \end{center}
\end{table}

The random baseline, $L_r$, is the expected value of $L$ under the null hypothesis.\footnote{Notice that $L$ is indeed the expected value of the length of a token but under a distinct setting (a distinct null hypothesis), where one picks a token uniformly at random over all tokens of a text and looks at its length. }
$L_r$ can be defined in more detail in two main equivalent ways:
\begin{enumerate}
    \item
    The value of $L$ that is expected if $L$ is recomputed after pairing the $f_i$'s and the $l_i$'s at random and recomputing $L$. The new value of $L$ depends on the variant of the null hypothesis. When shuffling the column for $f_i$ in the matrix (\autoref{tab:example}), the new $L$ is 
    \begin{equation*}
    L'= \frac{1}{T} \sum_{i = 1}^n f_i' l_i.
    \end{equation*}    
    When shuffling the column for $l_i$  and recomputing $L$, the new $L$ is 
    \begin{equation*}
    L' = \frac{1}{T} \sum_{i = 1}^n f_i l_i'.
    \end{equation*}        
    When shuffling both columns, the new $L$ is 
    \begin{equation*}
    L' = \frac{1}{T} \sum_{i = 1}^n f_i' l_i'.
    \end{equation*}
    \item 
    The average value of $L$ that is expected over all possible shufflings in one of the variants of the null hypothesis. 
    In the example in \autoref{tab:all_mappings}, on shuffling only the $l_i$ column, 
$$ L_r = \frac{\frac{155}{125} + \frac{170}{125} + \frac{235}{125} + \frac{265}{125} + \frac{330}{125} + \frac{345}{125}}{6} = \frac{155 + 170 + 235 + 265 + 330 + 345}{125 \cdot 6} = 2.$$
\end{enumerate}
We use $\expect[X]$ to refer to the expected value of a random variable $X$ under some variant of the null hypothesis above. Then $$L_r = \expect[L'],$$
where $L'$ is the value of $L$ resulting from some shuffling.

In quantitative linguistics, the mean length of tokens ($L$) is also known as dynamic word length \parencite{Chen2015a} and corresponds to the mean length of the words in a text. The mean length of types ($M$), defined as
$$M = \frac{1}{n} \sum_{i=1}^n l_i,$$
is also known as the static word length and corresponds to average length of the headwords in a dictionary \parencite{Chen2015a}.
Interestingly, the following property states that $L_r$ turns out to be $M$ independently of the variant of the null hypothesis under consideration.

\begin{property}
The expected value of $L'$ under any variant of the null hypothesis is $L_r = M$.
\label{prop:random_baseline}
\end{property}
\begin{proof}
We analyze $\expect[L']$ under each of the variants of the null hypothesis. 

{\em Dual shuffling.} Applying the linearity of expectation and independence between the shuffling of the  $p_i$ column of the that of the $l_i$ column, we obtain
\begin{eqnarray*}
\expect[L'_1] & = & \expect\left[\sum_{i=1}^n p_i' l_i' \right] \\
           & = & \sum_{i=1}^n \expect[p_i' l_i'] \\
           & = & \sum_{i=1}^n \expect[p_i'] \expect [l_i'].      
\end{eqnarray*}
Noting that
\begin{eqnarray*}
\expect[p_i'] & = & \frac{1}{n} \sum_{i=1}^n p_i = \frac{1}{n} \\
\expect[l_i'] & = & \frac{1}{n} \sum_{i=1}^m l_i = M,
\end{eqnarray*}
we finally obtain
\begin{equation}
\expect[L'] = \sum_{i=1}^n \frac{M}{n} = M.
\end{equation}

{\em Single shuffling of the $l_i$ column}. 
Applying the linearity of expectation and the fact that the column of $p_i$ remains constant, we obtain
\begin{eqnarray*}
\expect[L'_1] & = & \expect\left[\sum_{i=1}^n p_i l_i' \right] \\
           & = & \sum_{i=1}^n p_i \expect[l_i']. 
\end{eqnarray*}
Recalling $\expect[l_i'] = M$,
we finally obtain
\begin{equation}
\expect[L'] = M \sum_{i=1}^n p_i = M.
\end{equation}

{\em Single shuffling of the $p_i$ column}.  
Applying the linearity of expectation and the fact that the column of $l_i$ remains constant, we obtain
\begin{eqnarray*}
\expect[L'_1] & = & \expect\left[\sum_{i=1}^n p_i' l_i \right] \\
           & = & \sum_{i=1}^n \expect[p_i'] l_i. 
\end{eqnarray*}
Recalling $\expect[p_i'] = \frac{1}{n}$,
we finally obtain
\begin{equation}
\expect[L'] = \frac{1}{n} \sum_{i=1}^n l_i =  M.
\end{equation}

\end{proof}
The previous finding indicates that the random baseline for $L$ is equivalent to assuming that all word types are equally likely, namely, replacing each $p_i$ by $1/n$.

\begin{table}
  \caption{\label{tab:all_mappings} All the $3! = 6$ permutations of the column $l_i$ in \protect \autoref{tab:example} that can be produced. Each permutation is indicated with letters from A to F. 
  $L'$, the mean length of types in a shuffling, is shown at the bottom for each permutation. } 
  
  \begin{center} 
  \begin{tabular}{llllllll}
      &       & A     & B     & C     & D     & E     & F    \\
  $i$ & $f_i$ & $l_i'$ & $l_i'$ & $l_i'$ & $l_i'$ & $l_i'$ & $l_i'$\\
  \hline 
  1   &  100  & 1     & 1     & 2     & 2     & 3     & 3    \\         
  2   &  20   & 2     & 3     & 1     & 3     & 1     & 2    \\
  3   &  5    & 3     & 2     & 3     & 1     & 2     & 1    \\
  \hline  
  \\
      &  $L'$  & $\frac{155}{125}=1.24$   & $\frac{170}{125}=1.36$   & $\frac{235}{125}=1.88$   & $\frac{265}{125}=2.12$   & $\frac{330}{125}=2.64$   & $\frac{345}{125}=2.76$  \\ 
  \end{tabular}
  \end{center}
\end{table}



\section{Material}
\label{sec:material}


\subsection{General information about corpora and languages}

We investigate the relationship between the frequency of a word and its length in languages from two collections: Common Voice Forced Alignments (\autoref{CVFA}), hereafter CV, and Parallel Universal Dependencies (\autoref{PUD}), hereafter PUD. 

All the preprocessed files used to produce the results from the original collections are available in the repository of the article.\footnote{In the \textit{data} folder of \url{\repository}.}

PUD comprises 20 distinct languages from 7 linguistic families and 8 scripts (\autoref{tab:coll_summary_pud}).
CV comprises 46 languages from 14 linguistic families (we include 'Conlang', i.e. 'constructed languages', as a family for Esperanto and Interlingua) and 10 scripts (\autoref{tab:coll_summary_cv}).
Both PUD and CV are biased towards the Indo-European family and the Latin script. The typological information (language family) is obtained from Glottolog 4.6\footnote{\url{https://glottolog.org/}}. 
The writing systems are determined according to ISO-15924 codes\footnote{\url{https://unicode.org/iso15924/iso15924-codes.html}}. In \autoref{tab:coll_summary_pud} and \autoref{tab:coll_summary_cv}, we show the scripts using their standard English names. For example, most languages from the Indo-European family are written in Latin scripts. We also categorize Chinese Pinyin and Japanese Romaji as Latin scripts.


\begin{table}[H]
\centering
\caption{Summary of the main characteristics of the languages in the PUD collection. For each language, we show the linguistic family, the writing system (namely script name according to ISO-15924) and various numeric parameters: $A$, the observed alphabet size (number of distinct characters), $n$, the number of word types, and $T$, the number of word tokens.}
\label{tab:coll_summary_pud}
\begin{tabular}{lllrrr}
\hline
Language & Family & Script & $A$ & $n$ & $T$ \\ 
\hline
 Arabic & Afro-Asiatic & Arabic &  20 & 3309 & 11667 \\ 
  Indonesian & Austronesian & Latin &  23 & 4501 & 16702 \\ 
  Russian & Indo-European & Cyrillic &  23 & 4666 & 11749 \\ 
  Hindi & Indo-European & Devanagari &  44 & 4343 & 20071 \\ 
  Czech & Indo-European & Latin &  33 & 7073 & 15331 \\ 
  English & Indo-European & Latin &  25 & 5001 & 18028 \\ 
  French & Indo-European & Latin &  26 & 5214 & 20407 \\ 
  German & Indo-European & Latin &  28 & 6116 & 18331 \\ 
  Icelandic & Indo-European & Latin &  32 & 6035 & 16209 \\ 
  Italian & Indo-European & Latin &  24 & 5606 & 21266 \\ 
  Polish & Indo-European & Latin &  31 & 7188 & 15191 \\ 
  Portuguese & Indo-European & Latin &  38 & 5661 & 21855 \\ 
  Spanish & Indo-European & Latin &  32 & 5750 & 21067 \\ 
  Swedish & Indo-European & Latin &  25 & 5624 & 16378 \\ 
  Japanese & Japonic & Japanese & 1549 & 4852 & 24737 \\ 
  Japanese-strokes & Japonic & Japanese & 1549 & 4852 & 24737 \\ 
  Japanese-romaji & Japonic & Latin &  24 & 4849 & 24734 \\ 
  Korean & Koreanic & Hangul & 379 & 6218 & 12307 \\ 
  Thai & Kra-Dai & Thai &  50 & 3573 & 20860 \\ 
  Chinese & Sino-Tibetan & Han (Traditional variant) & 2038 & 4970 & 17845 \\ 
  Chinese-strokes & Sino-Tibetan & Han (Traditional variant) & 2038 & 4970 & 17845 \\ 
  Chinese-pinyin & Sino-Tibetan & Latin &  50 & 4970 & 17845 \\ 
  Turkish & Turkic & Latin &  28 & 6587 & 13799 \\ 
  Finnish & Uralic & Latin &  24 & 6938 & 12701 \\ 
   \hline

\end{tabular}
\end{table}

\begin{table}[H]
\centering
\caption{Summary of the main characteristics of the languages in the CV collection. For every language we show its linguistic family, the writing system (namely script name according to ISO-15924) and various numeric parameters: $A$, the observed alphabet size (number of distinct characters), $n$,  the number of word types, and, $T$, the number of word tokens.
'Conlang' stands for 'constructed language', that is an artificially created language. This is not a family in the proper sense as Conlang languages are not related in the common linguistic family sense.
} 
\label{tab:coll_summary_cv}
\begin{tabular}{lllrrr}
\hline
Language & Family & Script & $A$ & $n$ & $T$ \\ 
\hline
 Arabic & Afro-Asiatic & Arabic &  31 & 6397 & 45825 \\ 
  Maltese & Afro-Asiatic & Latin &  31 & 8058 & 44112 \\ 
  Vietnamese & Austroasiatic & Latin &  41 & 370 & 938 \\ 
  Indonesian & Austronesian & Latin &  22 & 3768 & 44210 \\ 
  Esperanto & Conlang & Latin &  27 & 27759 & 406261 \\ 
  Interlingua & Conlang & Latin &  20 & 5126 & 30504 \\ 
  Tamil & Dravidian & Tamil &  29 & 1210 & 6439 \\ 
  Persian & Indo-European & Arabic &  38 & 13115 & 1662508 \\ 
  Assamese & Indo-European & Assamese &  43 & 971 & 1813 \\ 
  Russian & Indo-European & Cyrillic &  32 & 31827 & 637686 \\ 
  Ukrainian & Indo-European & Cyrillic &  34 & 14337 & 120760 \\ 
  Panjabi & Indo-European & Devanagari &  37 &  84 &  98 \\ 
  Modern Greek & Indo-European & Greek &  33 & 5813 & 37880 \\ 
  Breton & Indo-European & Latin &  28 & 4228 & 38237 \\ 
  Catalan & Indo-European & Latin &  39 & 79112 & 3294206 \\ 
  Czech & Indo-European & Latin &  33 & 15518 & 147582 \\ 
  Dutch & Indo-European & Latin &  23 & 10225 & 316498 \\ 
  English & Indo-European & Latin &  28 & 173023 & 9828713 \\ 
  French & Indo-European & Latin &  49 & 160243 & 3729370 \\ 
  German & Indo-European & Latin &  30 & 148436 & 4230565 \\ 
  Irish & Indo-European & Latin &  23 & 2251 & 22593 \\ 
  Italian & Indo-European & Latin &  34 & 54996 & 811783 \\ 
  Latvian & Indo-European & Latin &  27 & 7251 & 29456 \\ 
  Polish & Indo-European & Latin &  32 & 25340 & 595411 \\ 
  Portuguese & Indo-European & Latin &  27 & 11509 & 283048 \\ 
  Romanian & Indo-European & Latin &  29 & 6423 & 33341 \\ 
  Romansh & Indo-European & Latin &  26 & 9614 & 43792 \\ 
  Slovenian & Indo-European & Latin &  24 & 5937 & 26304 \\ 
  Spanish & Indo-European & Latin &  33 & 75010 & 1842474 \\ 
  Swedish & Indo-European & Latin &  25 & 4371 & 62951 \\ 
  Welsh & Indo-European & Latin &  22 & 11143 & 539621 \\ 
  Western Frisian & Indo-European & Latin &  30 & 8383 & 63073 \\ 
  Oriya & Indo-European & Odia &  41 & 764 & 1700 \\ 
  Dhivehi & Indo-European & Thaana &  27 & 111 & 1284 \\ 
  Georgian & Kartvelian & Georgian &  25 & 6505 & 12958 \\ 
  Basque & Language isolate & Latin &  21 & 24748 & 458071 \\ 
  Mongolian & Mongolic & Mongolian &  31 & 14608 & 70217 \\ 
  Kinyarwanda & Niger-Congo & Latin &  26 & 133815 & 1939810 \\ 
  Abkhazian & Northwest Caucasian & Cyrillic &  28 & 119 & 156 \\ 
  Hakha Chin & Sino-Tibetan & Latin &  23 & 2499 & 17776 \\ 
  Chuvash & Turkic & Cyrillic &  22 & 4311 & 13583 \\ 
  Kirghiz & Turkic & Cyrillic &  30 & 10130 & 61844 \\ 
  Tatar & Turkic & Cyrillic &  34 & 21823 & 144356 \\ 
  Yakut & Turkic & Cyrillic &  28 & 7904 & 22577 \\ 
  Turkish & Turkic & Latin &  31 & 8926 & 107686 \\ 
  Estonian & Uralic & Latin &  23 & 28691 & 121549 \\ 
   \hline

\end{tabular}
\end{table}

\subsection{The datasets}

We measure word length in two main ways: \textit{duration in time} and \textit{length in characters}. Concerning Chinese and Japanese, we additionally consider the number of strokes and the number of characters of their romanization (i.e. Pinyin for Chinese and Romaji for Japanese). 

 
 Given these datasets, word durations are obtained only from CV. Word lengths in characters are obtained from both CV as well as from PUD. Word lengths in strokes, and word lengths in characters after romanization, are obtained only from PUD.

\subsubsection{Common Voice Forced Alignments} \label{CVFA}

The Common Voice Corpus\footnote{\url{https://commonvoice.mozilla.org/en/datasets}} is an open source dataset of recorded voices uttering sentences in many different languages. The amount of data, as well as the source and topic of each sentence, depends considerably on the language and the corpus version. Specifically, the Common Voice Corpus 5.1 contains information on 54 languages and dialects.

Common Voice Forced Alignments (CVFA)\footnote{\url{https://github.com/JRMeyer/common-voice-forced-alignments}} were created by Josh Meyer using the Montreal Forced Aligner\footnote{\url{https://github.com/MontrealCorpusTools/Montreal-Forced-Aligner}} on top of the Common Voice Corpus 5.1. Kabyle, Upper Sorbian and Votic were left out of the alignments for an undocumented reason. Therefore, CVFA contains information on 51 languages.

In our analyses, Japanese and the three Chinese dialects were excluded as the forced aligner failed to correctly extract words from sentences. In addition, both Romansh dialects were fused into a single Romansh language. 
Indeed, given the nature of this corpus, all languages are likely to be represented by more than one dialect.

Notice that Abkhazian, Panjabi, and Vietnamese have a critically low number of tokens ($T<1000$ in \autoref{tab:coll_summary_cv}). However, we decided to include them in the analyses so as to understand their limitations related to corpus size.

\subsubsection{Parallel Universal Dependencies} \label{PUD}

The Universal Dependencies (UD)\footnote{\url{https://universaldependencies.org/}} collection is an open source dataset of annotated sentences, in which the amount of data depends on each language. The Parallel Universal Dependencies (PUD) collection is a parallel subset of 20 languages from the UD collection, consisting of 1000 sentences. 
It allows for a cross-language comparison, controlling for content and annotation style.

In \autoref{tab:coll_summary_pud}, we show the characteristics of the languages in PUD. For traditional Chinese and Japanese, we also include word lengths in romanizations (Pinyin and Romaji respectively), as well as word lengths measured in strokes,
resulting in a total of 24 language files. 
Notice that three Japanese words that are hapax legomena  could not be romanized and thus the number of tokens and types varies slightly with respect to the original Japanese characters (\autoref{tab:coll_summary_pud}). 


\section{Methodology}
\label{sec:methodology}

All the code used to produce the results is available in the repository of the article.\footnote{In the \textit{code} folder of \url{\repository}.}

\subsection{The units of length}

\subsubsection{Duration}

The duration of a word for a given language is estimated by computing the median duration in seconds across all its occurrences in utterances in the CV corpus. 
All words with equal orthographic form are assumed to be the same type. The median is preferred over the mean as it is less sensitive to outliers (that may be produced by forced alignment errors) and better suited to deal with heavy-tailed distributions \parencite{Hernandez2019a}. Given the oral nature of the data, we do expect to observe some variation in the duration of words, 
due to differences between individuals, and variation within a single individual. This is more generally in line with speakers acting as complex dynamical systems \parencite{Kello2010a}. For these reasons, median duration is preferred for research on the law of abbreviation in acoustic units \parencite{Torre2019a,Watson2020a}. 

\subsubsection{Length in characters}

Word length in characters is measured by counting every Unicode UTF-8 character present in a word. Special characters such as ``='' were removed. Characters with stress accents are considered as different from their non-stressed counterpart (e.g. ``a'' and ``à'' are considered separate characters). Following best practices from \parencite{Meylan2021a}, characters were always kept in UTF-8. 

\subsubsection{Length in strokes}

Japanese Kanji and Chinese Hanzi were turned into strokes using the \textit{cihai} Python library.\footnote{\url{https://github.com/cihai/cihai}} 
In Japanese characters other than Kanji, namely Japanese Kana, the number of strokes in printed versus hand-written modality can differ (Chinese Hanzi and Japanese Kanji have the same number of strokes in printed version or hand-written version). Here we counted the number of strokes in printed form.
Japanese Kana were converted into printed strokes by using a hand-crafted correspondence table, since Kana is not part of the CJK unified character system.
This table was created by us and checked by a native linguist (S. Komori from Chubu University, Japan). It is available in the repository of the article.\footnote{In the \textit{data/other} folder of \url{\repository}.}

In case of discrepancies on the number of strokes for a given character, the most typical printed version was chosen.

\subsubsection{Length in Pinyin and Romaji}

\begin{CJK*}{UTF8}{ipxm}
Chinese Pinyin was obtained using the \textit{cihai} package as above, while the Japanese Romaji was obtained with the \textit{cutlet} Python library.\footnote{\url{https://github.com/polm/cutlet}} The latter uses Kunrei-shiki  romanization (since it is the one used officially by the government of Japan) and the spelling of foreign words 
is obtained in its native reading
(e.g. ``カレー'' is romanized as ``karee'' instead of ``curry''). There are some particularities with the  romanization of Kanji characters by \textit{cutlet}. For example, in the case of the word ``year'' (年), it chose the reading of ``Nen'' instead of ``Tosi'', which would be the expected one.

A more systematic issue with Japanese romanization is that it does not provide means to indicate pitch accents, which are implicitly present in Kanji. For example, ``日本'' ``Ni$\uparrow$hon'' (``Japan'') is romanized as simply ``Nihon''. Therefore, the alphabet size of romanized Japanese is smaller than it should be, compared to other languages where, as stated before, stress accents are counted as distinctive features of characters.
\end{CJK*}

\subsection{Tokenization} 

Tokenization is already given in each dataset and we borrow it for our analyses. 
Thus tokenization methods are not uniform for CV and PUD and are not guaranteed to be uniform among languages even within each of these datasets.

\subsection{Filtering of tokens}

Examining our datasets, we noticed that in some text files there was a considerable number of unusual character strings, as well as foreign words (written in different scripts). These need to be filtered out in order to obtain a ``clean'' set of word types.
To this end we filter out tokens following a two step procedure:
\begin{enumerate}
    \item 
    {\em Mandatory elementary filtering}. This filter consists of:
    \begin{itemize}
    \item 
    {\em Common filtering}. In essence, it consists of the original tokenization and the removal of tokens containing digits. In each collection, the original tokenizer yields tokens that may contain certain punctuation marks. 
Due to the nature of the CV dataset, the bulk of punctuation was already removed via the Montreal Forced Aligner with some exceptions. For instance, single quotation (in particular ``''') is a punctuation sign that is kept within a word token in CV, as it is necessary for the formation of clitics in multiple languages, such as in English or French. In PUD, as a part of UD, contractions are split into two word types. ``can't'' is split into ``ca'' ``n't'' (in CV ``can't'' would remain as just one token). In both collections, words containing ASCII digits are removed because they do not reflect phonemic length and can be seen as another writing system.

    \item
    {\em Specific filtering.} In case of the PUD collection, we excluded all tokens with Part-of-Speech (POS) tag `PUNCT'. In case of the CV collection, we removed tokens tagged as <unk> or null tokens, namely tokens that either could not be read or that represent pauses.
    \item
    {\em Lowercasing.} Every character is lowercased. In the case of CV, this is already given by the Montreal Forced Aligner, while in the case of PUD, tokens are lowercased by means of the {\em spaCy} Python package.\footnote{\url{https://spacy.io/}}
    \end{itemize}
    
    
    \item
    {\em Optional filtering}. This is a new method that is applied after the previous filter and described in \autoref{sec:filter}. 
\end{enumerate}

\subsection{A new method to filter out unusual characters}
\label{sec:filter}

It has been pointed out that ``chunk'' words and loanwords can distort the results of quantitative analyses of word lengths \parencite{Meylan2021a}. Indeed, especially the files of the Common Voice Corpus feature a considerable number of word tokens which do not consist of characters belonging to the primary alphabet of the respective writing system. \textcite{Meylan2021a} proposed to use dictionaries to exclude such anomalous words. However, this is not feasible for our multilingual datasets, as loanword dictionaries are not available for this large number of diverse languages (\autoref{tab:coll_summary_pud} and \autoref{tab:coll_summary_cv}). The Intercontinental Dictionary Series,\footnote{\url{https://ids.clld.org/}} for example, contains only around half of the languages in our analysis, so it is not applicable to many of them. Hence, this approach would lead to a non-uniform treatment of different languages and texts. Selecting a matched set of semantic concepts across languages using a lexical database is also infeasible due to similar reasons. 

Against this backdrop, we decided to develop an unsupervised method to filter out words which contain highly unusual characters. For a given language, the method starts 
by assuming that the strings (after the mandatory filtering illustrated above) contain characters of two types: characters of the working/primary alphabet as well as other characters. We hypothesize that the latter are much less frequent than the former. Following this rationale, we apply the $k$-means algorithm of the {\em Ckmeans R} package\footnote{\url{https://cran.r-project.org/web/packages/Ckmeans.1d.dp/index.html}} to split the set of characters into the two groups based on the logarithm of the frequency of the characters.\footnote{The motivation for taking logarithms of frequencies is three-fold: First, this brings observations closer together. Note that the $k$-means algorithm prefers high-density areas. Second, this transforms the frequencies into a measure of surprisal, following standard information theory \parencite{Shannon1948}. Third, manual inspection suggests that the logarithmic transformation is required to produce an accurate split.}
To maximize the power of the clustering method, we use the exact method with $k=2$ for one dimension instead of the customary approximate method.
We then keep the high frequency cluster as the real working alphabet and filter out the word tokens that contain characters not belonging to this high frequency cluster.




We illustrate the power of the method by showing working alphabets that are obtained on CV, that is the noisiest one of the collections. 

In English, the working alphabet is defined by the 26 English letters and quotation marks (``''', ``’''). These quotation marks are used often in clitics, and as such are correctly identified as part of the encoding, since, for example, ``can't'' and ``cant'' are different words in meaning, with ``can't'' meaning ``can not'', while ``cant'' is a statement on a religious or moral subject that is not believed by the person making the statement, with the differentiating feature being the ``'''. Therefore, the working alphabet becomes 5 vowels (``a'', ``e'', ``i'', ``o'', ``u''), 21 consonants (``b'', ``c'', ``d'', ``f'', ``g'', ``h'', ``j'', ``k'', ``l'', ``m'', ``n'', ``p'', ``q'', ``r'', ``s'', ``t'', ``v'', ``w'', ``x'', ``y'', ``z'') and 2 kinds of quotation marks (``''', ``’'').


In Russian, the working alphabet comprises 9 vowels ( ``\foreignlanguage{russian}{а}'',   ``\foreignlanguage{russian}{о}'', ``\foreignlanguage{russian}{у}'', ``\foreignlanguage{russian}{ы}'', ``\foreignlanguage{russian}{э}'',  ``\foreignlanguage{russian}{я}'', ``\foreignlanguage{russian}{ю}'', ``\foreignlanguage{russian}{и}'', ``\foreignlanguage{russian}{е}''),
a semivowel / consonant 
``\foreignlanguage{russian}{й}'',
20 consonants ( 
``\foreignlanguage{russian}{б}'', ``\foreignlanguage{russian}{в}'',  ``\foreignlanguage{russian}{г}'', ``\foreignlanguage{russian}{д}'',  ``\foreignlanguage{russian}{ж}'', ``\foreignlanguage{russian}{з}'',
``\foreignlanguage{russian}{к}'', ``\foreignlanguage{russian}{л}'', ``\foreignlanguage{russian}{м}'', ``\foreignlanguage{russian}{н}'', ``\foreignlanguage{russian}{п}'', ``\foreignlanguage{russian}{р}'', ``\foreignlanguage{russian}{с}'', ``\foreignlanguage{russian}{т}'', ``\foreignlanguage{russian}{ф}'', ``\foreignlanguage{russian}{х}'', ``\foreignlanguage{russian}{ц}'', ``\foreignlanguage{russian}{ч}'', ``\foreignlanguage{russian}{ш}'', ``\foreignlanguage{russian}{щ}'')
and 2 modifier letters 
(``\foreignlanguage{russian}{ъ}'', ``\foreignlanguage{russian}{ь}'').

In Italian, it comprises 5 vowels (``a'', ``e'', ``i'', ``o'', ``u''), 21 consonants (``b'', ``c'', ``d'', ``f'', ``g'', ``h'', ``j'', ``k'', ``l'', ``m'', ``n'', ``p'', ``q'', ``r'', ``s'', ``t'', ``v'', ``w'', ``x'' , ``y'', ``z'') and 6 instances of the 5 vowels containing a diacritic mark (``à'', ``è'', ``é'', ``ì'', ``ò'', ``ù''). 

The unsupervised filter method filter is not applied to Chinese, Japanese and Korean as, given their nature, this would exclude letters that actually belong to the real alphabet. 
In \autoref{app:no_filter} we analyze the impact of the optional filter and provide arguments for not applying the unsupervised filter to these languages. 
As a compensation, strings that contain non-CJK characters are filtered out in Chinese and Japanese as a part of the optional filter. In Korean, only a few characters are not proper Hangul and thus such a complementary filtering is not necessary.

\subsection{Immediate constituents in writing systems}
\label{sec:immediate_constituents}

When measuring word length in written languages, we are using \textit{immediate constituents} of written words. In Romance languages, the immediate constituents are letters of the alphabet, which are a proxy for phonemes. For syllabic writing systems (as Chinese in our dataset), these are characters that correspond to syllables. 
In addition, for Chinese and Japanese,  we are considering two other possible units for word length, which are not immediate constituents, but alternative ways of measuring word lengths which could provide useful insights: strokes and letters in Latin script romanizations. That means that for each of these languages words are unfolded into three systems, one for each unit of encoding (original characters, strokes, romanized letters/characters). In the hierarchy from words to other units, only the original characters are immediate constituents. 

\subsection{Statistical testing}

\subsubsection{Correlation}

When measuring the association between two variables, we  use both Pearson correlation and Kendall correlation \parencite{Conover1999a}. 
Note that the traditional view of Pearson correlation as a measure of linear association and thus not suitable for non-linear association has been challenged \parencite{vandenJeuvel2022a}. 

\subsubsection{How to test for the law of abbreviation}

We used a left-sided correlation test to verify the presence of the law of abbreviation. In a purely exploratory or atheoretic exploration, one should use a two-sided test. In an exploration guided by theory, namely regarding the law of abbreviation as a manifestation of compression, the test should be left-sided as theory predicts that $\tau(p,l)$ cannot be positive in case of optimal coding \parencite{Ferrer2019c}. 

\subsubsection{How to test for compression}

In the context of the null hypothesis of a random mapping of type probabilities into type lengths, testing that compression (minimization of $L$) has some effect on actual word lengths is easy because $L$ is a linear function of $r$, the Pearson correlation between word length and word probability (Appendix \ref{app:theory}). In particular,
$$L = a r + L_r,$$
where $a = (n-1)s_p s_l$, being $n$ the number of types and $s_p$ and $s_l$, respectively,the standard deviation of type probabilities and type lengths.
In such random mappings, $L_r$, $s_p$ and $s_l$ remain constant and then testing if $r$ is significantly small is equivalent to testing if $L$ is significantly small (notice $a \geq 0$).

\subsubsection{Controlling for multiple testing}

When performing multiple correlation tests at the same time, it becomes easier to reject the null hypothesis simply by chance. To address this problem we used a Holm-Bonferroni correction to \textit{p}-values.\footnote{\url{https://stat.ethz.ch/R-manual/R-devel/library/stats/html/p.adjust.html}} We applied the correction when checking the law of abbreviation in the languages of a collection, so as to exclude the possibility that the law of abbreviation is found many times simply because we are testing it in many languages.

\section{Results}
\label{sec:results}

In \autoref{sec:introduction}, we highlighted the importance of distinguishing 
between direct and indirect evidence of compression. Against this theoretical backdrop, here we first investigate the presence of Zipf's law of abbreviation in languages. Then we investigate direct evidence of compression with the help of the new random baseline. 

\subsection{The law of abbreviation revisited}

\label{sec:law_of_abbrebiation}

\begin{figure}[H]
    \centering
    \begin{subfigure}{.45\textwidth}
        \caption{}
        \includegraphics[width=8cm, height=10cm]{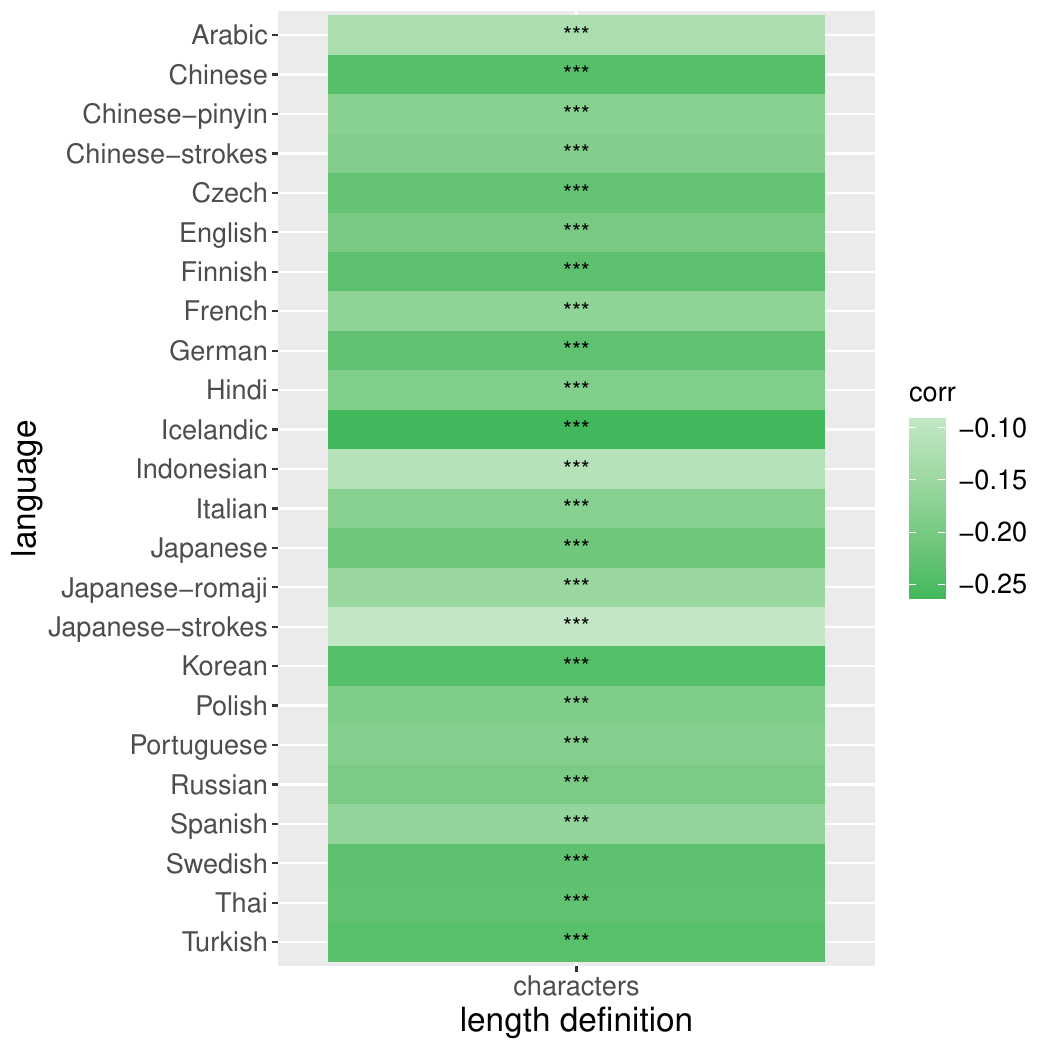}
    \end{subfigure}
    \hskip\baselineskip
    \begin{subfigure}{0.45\textwidth}
        \caption{}
        \includegraphics[width=8cm, height=10cm]{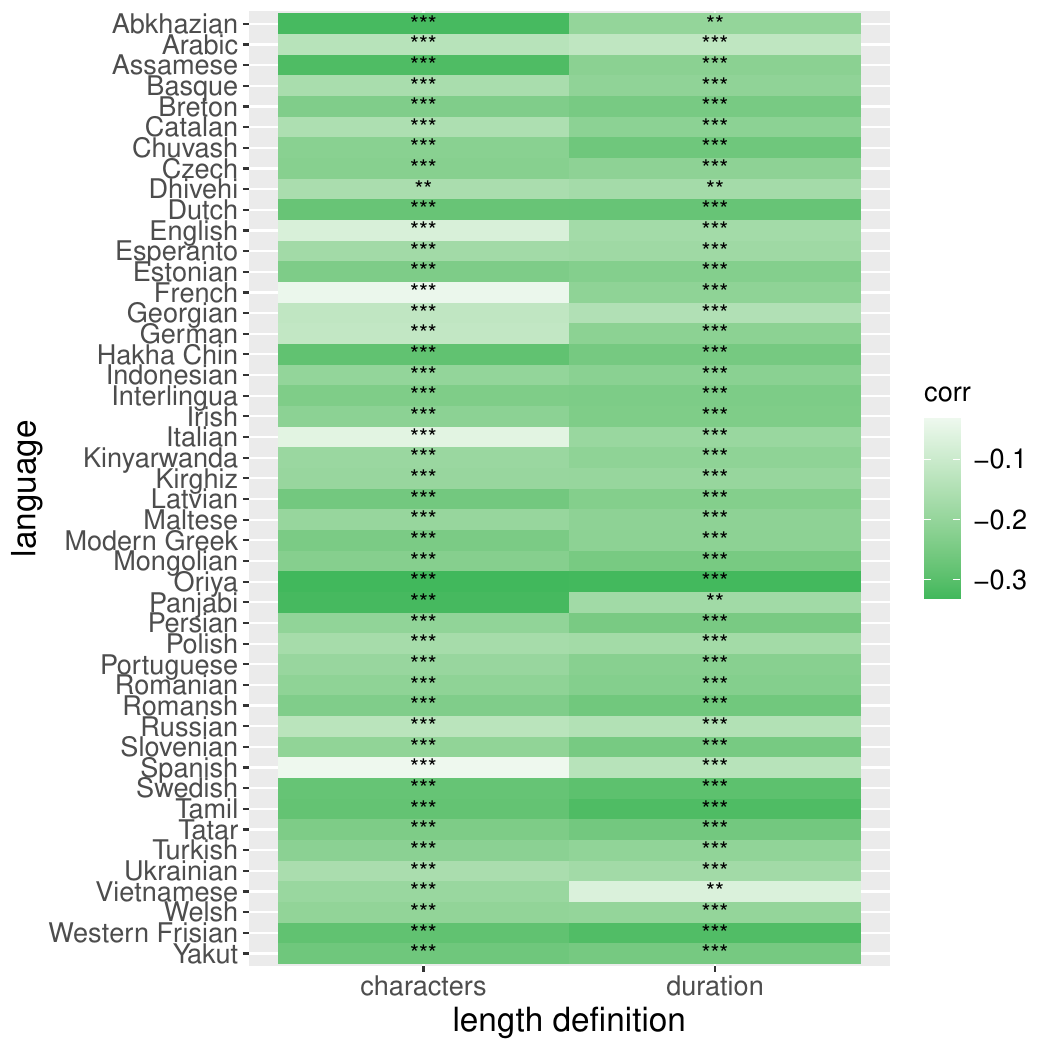}
    \end{subfigure}
        \begin{subfigure}{.45\textwidth}
        \caption{}
        \includegraphics[width=8cm, height=10cm]{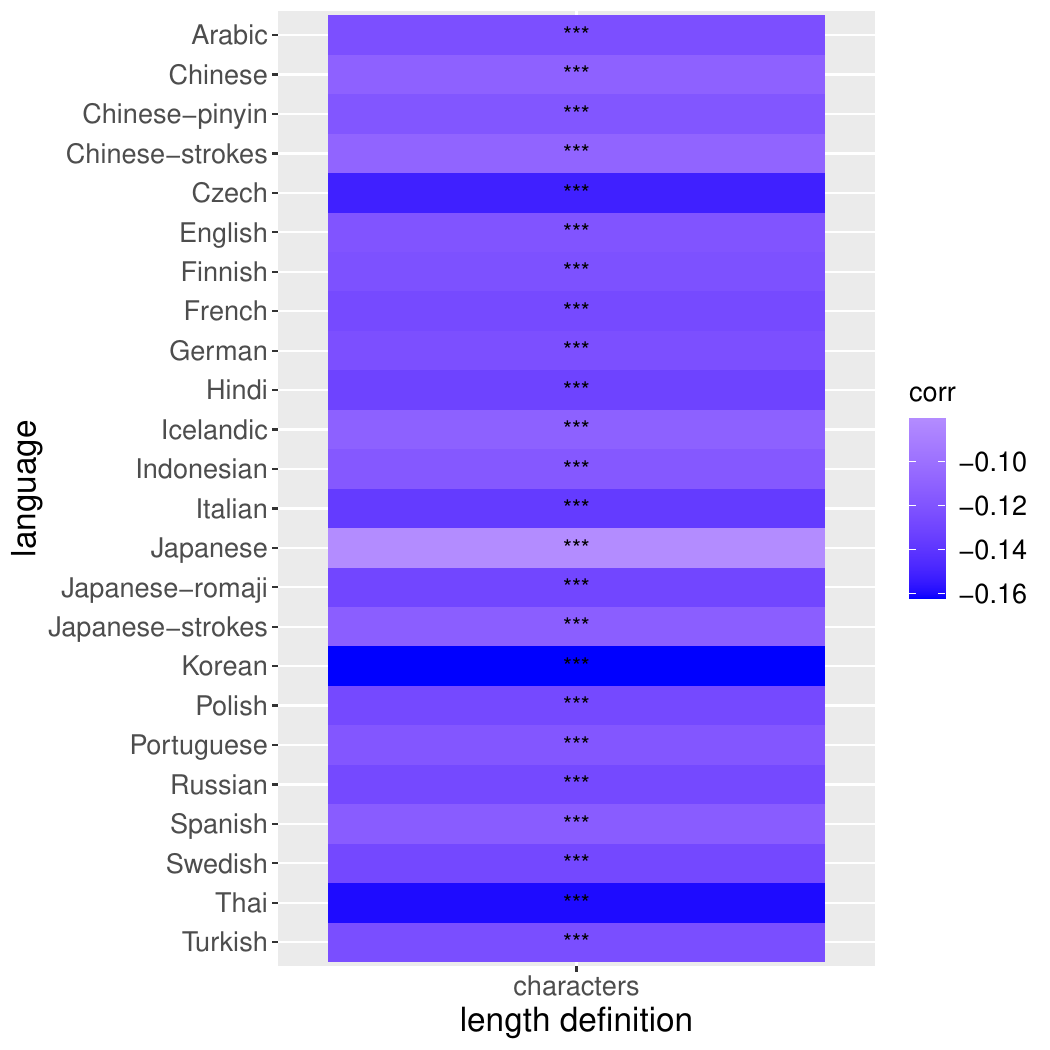}
    \end{subfigure}
    \hskip\baselineskip
    \begin{subfigure}{0.45\textwidth}
        \caption{}
        \includegraphics[width=8cm, height=10cm]{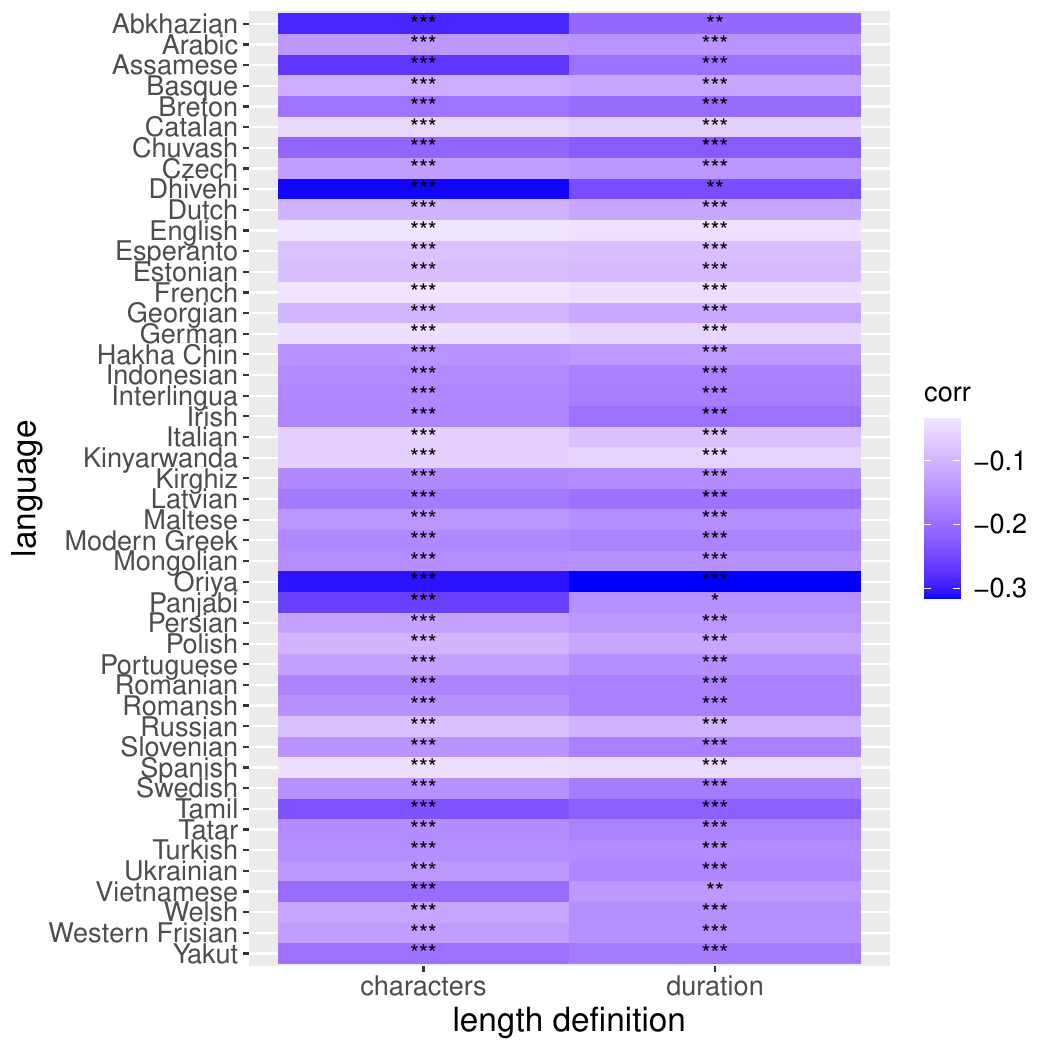}
    \end{subfigure}
    \caption{\label{fig:corr_significance} 
     The correlation between frequency and length across languages.
     '***' indicates a Holm-Bonferroni corrected $p$-value lower than or equal to 0.01, '**' indicates lower than or equal to 0.05 but smaller than 0.1 and '*' indicates lower than or equal to 0.1. Here '*' symbols are not used to indicate significance but p-value ranges. 
     (a) Kendall $\tau$ correlation in PUD (word length in characters). (b) Kendall $\tau$ correlation in CV 
     (left: word length in characters; right: word length in duration).
     (c) Same as (a) with Pearson $r$ correlation. (d) Same as (b) with Pearson $r$ correlation. 
     }
\end{figure}

We investigate the presence of the law of abbreviation by means of left-sided correlation tests for the association between frequency and length. We use both Kendall correlation, as suggested by theory on the origins of the law \parencite{Ferrer2019c}, and Pearson's. 
For each language, we show the significance level of the relationship, color-coded by the value of the correlation coefficient. \autoref{fig:corr_significance} (a,b) indicates that the law holds in all languages -- regardless of the definition of word length -- 
when Kendall $\tau$ correlation is used. 
In both collections, we find Kendall $\tau$ correlation coefficients significant at the 99\% confidence level, except for Dhivehi in the CV collection when length is measured in characters, and Abkhazian, Dhivehi, Panjabi and Vietnamese when length is measured in duration. However, note that these are all still significant at the 95\% confidence level. When Pearson correlation is used instead, \autoref{fig:corr_significance} (c) shows that the picture remains the same in PUD.
The main findings are the same also in CV (\autoref{fig:corr_significance} (d)),
 but when length is measured in duration Panjabi ceases to be significant at the 95\% confidence level. Overall, we only fail to find the law of abbreviation in Panjabi given word durations, and using Pearson correlation. This is most probably related to undersampling, as this particular language only features 98 tokens (\autoref{tab:coll_summary_cv}). 

\subsection{Real word lengths versus the random baseline}

\begin{figure}[H]
    \centering
    \begin{subfigure}{.4\textwidth}
        \caption{}
        \includegraphics[scale=0.4]{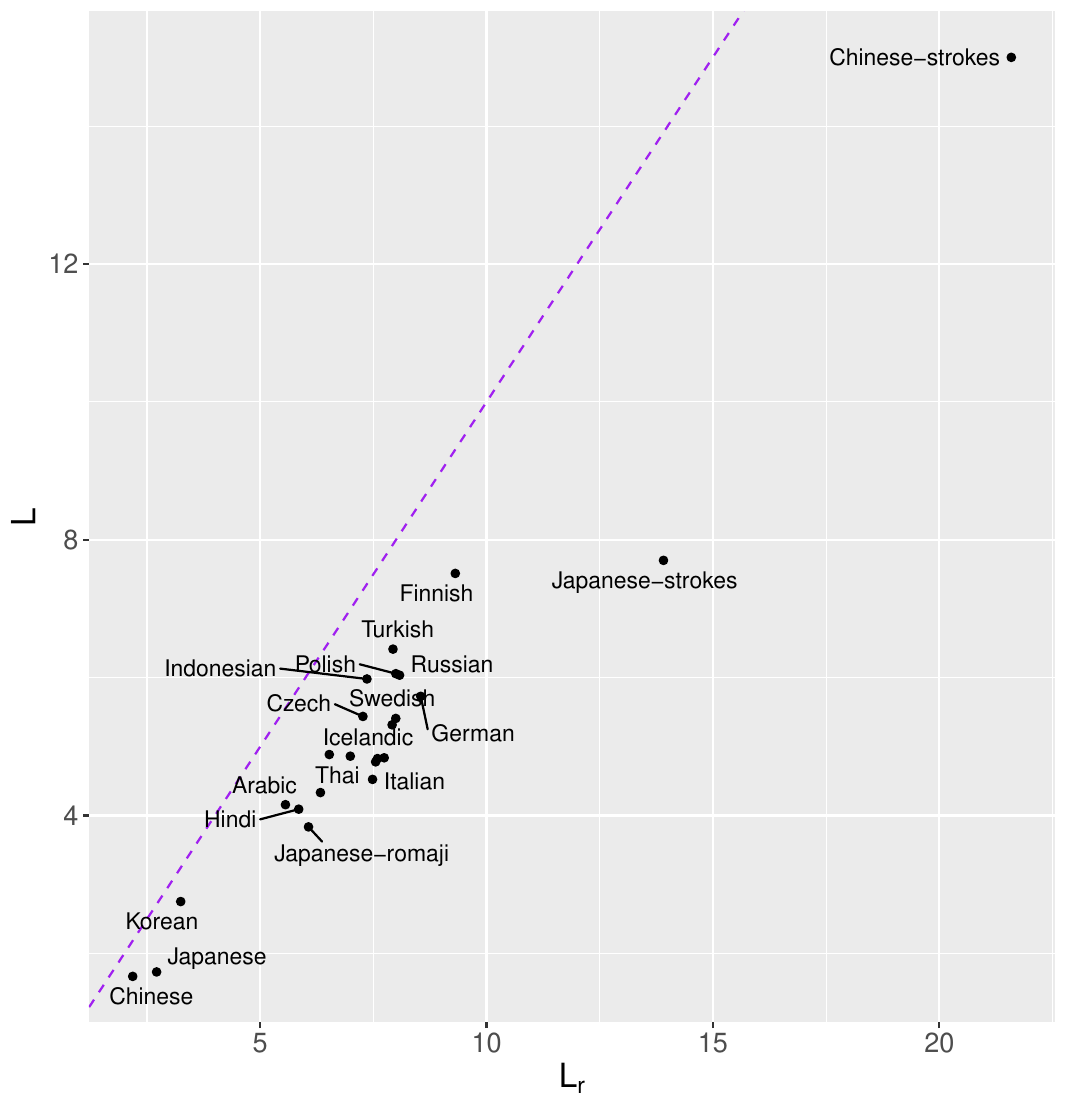}
    \end{subfigure}
    \hskip\baselineskip
    \begin{subfigure}{0.4\textwidth}
        \caption{}
        \includegraphics[scale=0.4]{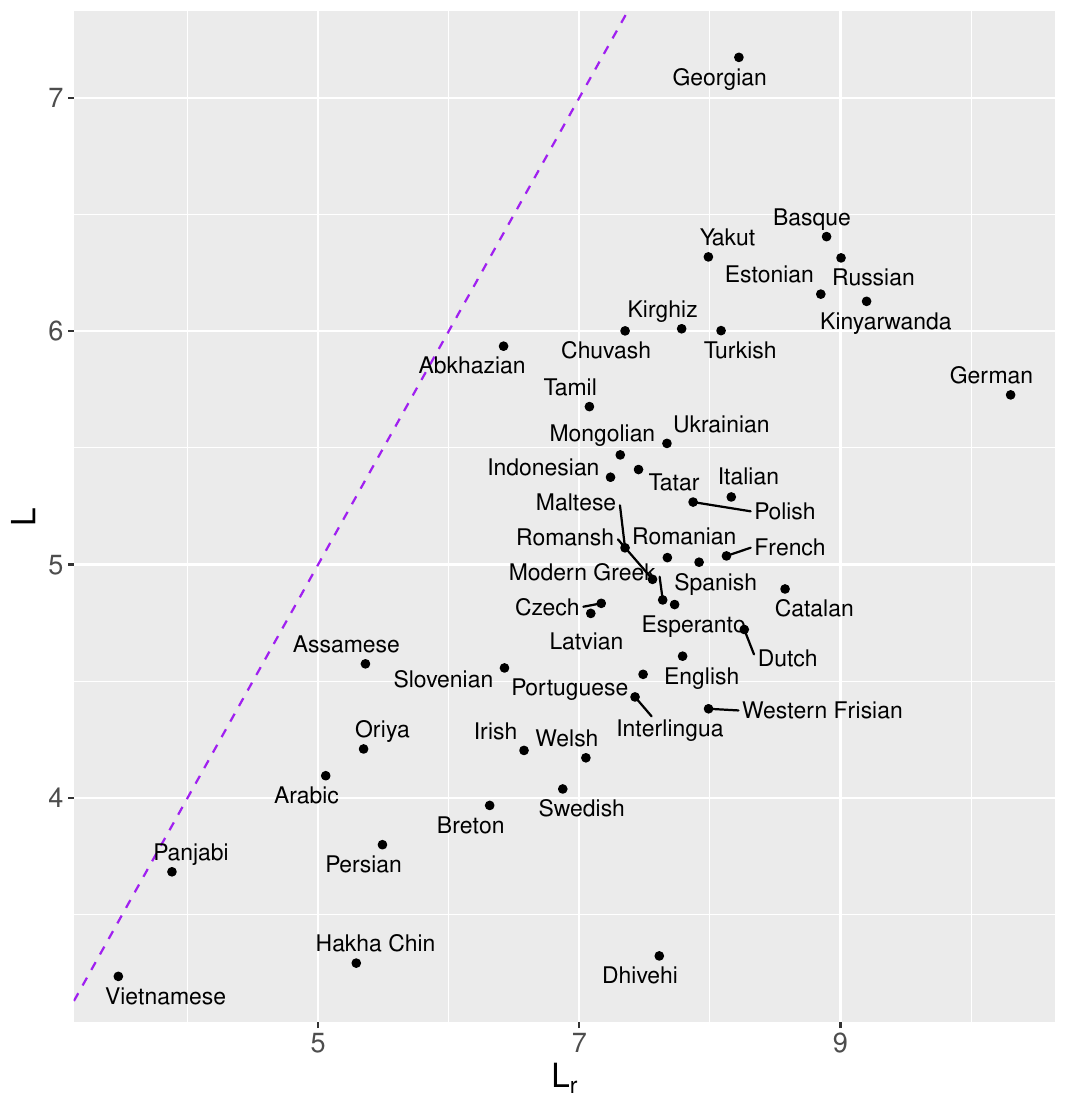}
    \end{subfigure}
        \begin{subfigure}{.45\textwidth}
        \caption{}
        \includegraphics[scale=0.4]{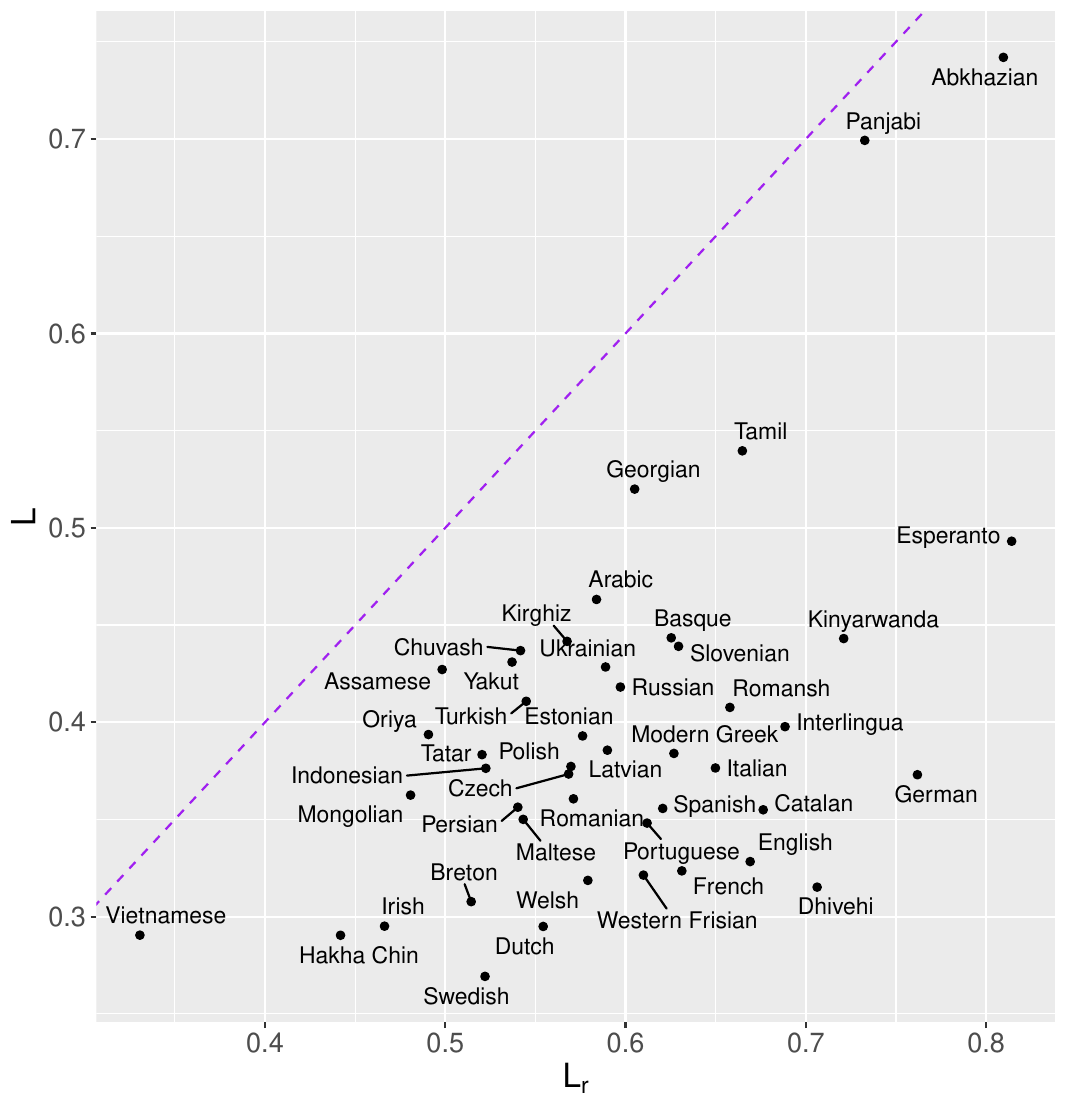}
    \end{subfigure}
    \caption{\label{fig:mean_word_length_versus_random_baseline} Mean word length ($L$) as a function of the random baseline ($L_r$) in languages. Every point stands for a language. The diagonal (long dashed line) indicates the line $L=L_r$. Languages with $L < L_r$ are located below the diagonal. (a) Languages in PUD with word length measured in characters (or strokes for Chinese and Japanese). (b) Languages in CV with word length measured in characters. (c) Languages in CV with word length measured in duration (seconds). 
    }
\end{figure}

We investigate the relationship between the actual mean word length ($L$) and the random baseline ($L_r$). We find that $L < L_r$ for all languages in every collection (\autoref{fig:mean_word_length_versus_random_baseline} and Tables \ref{tab:opt_scores_pud}, \ref{tab:opt_scores_cv_characters}, \ref{tab:opt_scores_cv_meadianDuration}). Interestingly, there is a large gap between $L$ and $L_r$ in the majority of languages, which is more compelling in CV with word durations (\autoref{fig:mean_word_length_versus_random_baseline}).  
Exceptions to the large gap -- as in the case of Panjabi and Abkhazian when length is measured in duration -- mainly concern languages with reduced sample sizes. The result holds even when alternative units of measurement are considered for Chinese and Japanese.

\autoref{fig:mean_word_length_versus_random_baseline} is reminiscent of Figure 4 of \textcite{Pimentel2021a} but our setting is much simpler (it only involves $L$ and $L_r$).

\subsection{Impact of disabling the filter of words that contain ``foreign'' characters }

All results presented in this section have been obtained after applying the new method to filter out highly unusual characters and words described in \autoref{sec:filter}. If the filter is disabled, we obtain some slight changes in the values, but the qualitative results summarized above remain the same.

\section{Discussion}

\label{sec:discussion}

\subsection{The universality of Zipf's law}

The first step of our analysis consisted in checking the universality of the law of abbreviation in the languages of our samples through a Kendall $\tau$ correlation test. Here, we introduced two methodological improvements with respect to previous research: using the Bonferroni-Holm correction for $p$-values, as well as word length in time given spoken utterances, rather than just characters in written form \parencite{Bentz2016a}. We also computed Pearson correlations for two reasons: (a) to verify the robustness of the conclusions and (b) to check the significance of the gap between $L$ and $L_r$ (the case of (b) is addressed in the next subsection). We find that the law of abbreviation holds in nearly all languages in our sample at a 95\% confidence level, independently from how word length is measured, and even after controlling for multiple testing. The only exception is Panjabi in CV, but only when length is measured in duration and Pearson $r$ correlation is used. Panjabi is also the language suffering most from under-sampling (only 98 tokens). 
Therefore, Panjabi cannot be considered a true exception to the law of abbreviation. 

Given the rather scarce evidence of the law of abbreviation in word durations in human language \parencite{Torre2019a}, we have taken step forward by providing evidence of it in 46 languages from 14 linguistic families. The massive agreement of the law of abbreviation even when orthographic word lengths are replaced by word durations in human languages provides stronger support for the law of abbreviation as a potentially universal pattern of human languages with respect to previous research relying on word length in characters \parencite{Bentz2016a} and often on a small number of linguistic families \parencite{Piantadosi2011a,Levshina2022a,Meylan2021a, Koplenig2022a}.


\subsection{Direct evidence of compression}

We have found that word lengths are shorter than expected by chance ($L < L_r$) in all languages in every collection (\autoref{fig:mean_word_length_versus_random_baseline}). Such a systematic finding is unlikely to be accidental and strongly indicates that compression is acting in all languages in our sample. Crucially, the finding holds independently of how word length is measured.
The ample evidence of compression even when orthographic word lengths are replaced by word durations in human languages provides stronger support for compression as a universal principle of the organization of languages with respect to previous research relying on word length in characters \parencite{Ferrer2012d}.

It could be argued that these findings constitute evidence of compression in ensembles of language but not in individual languages. The reason is that $L < L_r$ does not imply that the difference between the actual word length and the random baseline is statistically significant for a single language. However, we have shown that the Pearson correlation is indeed a linear function of $L$ and $L_r$ (Appendix \ref{app:theory}) and thus $L$ is significantly small in every language where the law of abbreviation has been confirmed using a Pearson correlation test. 


Finally, the direct correspondence we have established between the average length of types ($M$) and the random baseline sheds new light on previous research. For instance, it has been shown that $M < L$ in Chinese characters in six time periods spanning two millennia \parencite[Fig. 4]{Chen2015a}, which now can be reinterpreted as a sign of compression of word lengths in Chinese in light of our theoretical findings. 

\subsection*{Future research}

In this article, we have introduced a new random baseline and unveiled a systematic gap between that random baseline and real mean word lengths that we have interpreted as direct evidence of compression. \autoref{fig:mean_word_length_versus_random_baseline} suggests that the gap is wider when word lengths are measured in duration rather than in characters. 
However, we have not quantified the magnitude of that gap and we have neither taken into consideration the gap between actual mean words lengths and the minimum baseline, that would be defined as the minimum word length that could be achieved under certain constraints \parencite{Cover2006a, Ferrer2019c, Pimentel2021a}. Future research should quantify the first gap in relation to the minimum baseline. As the random baseline is crucial to asses the degree of optimality of word lengths, we have paved the way for exploring the degree of optimality of word lengths in characters or duration in languages.

\iftoggle{anonymous}{}{
\section*{Authors' contributions}
SP: Conceptualization, Formal Analysis, Investigation, Software, Supervision, Validation, Visualization, Writing-original draft, Writing-review \& editing;
ACM: Data curation, Resources, Software;
JCM: Writing-review \& editing, Resources;
MW: Writing-review \& editing;
CB: Conceptualization, Writing-review \& editing;
RFC: Conceptualization, Formal analysis, Funding acquisition, Methodology, Project Administration, Supervision, Writing-original draft, Writing-review \& editing.
}

\section*{Acknowledgments}

This article is one of the products of the research project of the 1st edition of the master degree course ``Introduction to Quantitative Linguistics'' at Universitat Politècnica de Catalunya. We are specially grateful to two students of that course: L. Alemany-Puig for helpful discussions and computational support, and M. Michaux for comments on early versions of the manuscript. We thank M. Farrús and A. Hernández-Fernández for advice on voice datasets. We also thank S. Komori for advice on Japanese, Y. M. Oh for advice on Korean and S. Semple for helping us to improve English. Last but not least, we are very grateful to an anonymous reviewer for very helpful comments.

SP is funded by the grant 'Thesis abroad 2021/2022' from the University of Milan. CB was partly funded by the \textit{Deutsche Forschungsgemeinschaft} (FOR 2237:  Words, Bones, Genes, Tools - Tracking Linguistic, Cultural and Biological Trajectories of the Human Past), and the \textit{Schweizerischer Nationalfonds zur Förderung der
Wissenschaftlichen Forschung} (Non-randomness in Morphological Diversity: A Computational Approach Based on Multilingual Corpora, 176305). RFC is supported by a recognition 2021SGR-Cat (01266 LQMC) from AGAUR (Generalitat de Catalunya).



\printbibliography



\begin{appendices} 

\counterwithin{figure}{section}
\counterwithin{table}{section}
\renewcommand\thefigure{\thesection\arabic{figure}}
\renewcommand\thetable{\thesection\arabic{table}}

\section{Theory}
\label{app:theory}



Here we review the relationship between \texorpdfstring{$L$}{L}, \texorpdfstring{$L_r$}{Lr} and Pearson correlation

Given two random variables $x$ and $y$ and a sample of $n$ points, $\left\{(x_1, y_1),...,(x_i, y_i),...,(x_n,y_n) \right\}$,
the sample covariance is defined as 
$$s_{xy} = \frac{1}{n-1}\left(\sum_{i=1}^n x_i y_i - n \bar{x}\bar{y}\right),$$
where $\bar{x}$ is the sample mean of $x$ and $\bar{y}$ is the sample mean for $y$, i.e. 
\begin{eqnarray*}
\bar{x}= \frac{1}{n} \sum_{i=1}^n x_i \\
\bar{y}= \frac{1}{n} \sum_{i=1}^n y_i.
\end{eqnarray*}
Now consider than the random variables are $p$ (the probability of a type) and $l$ (the length/duration of a type) instead of $x$ and $y$. 
Then our sample of $n$ points is $\left\{(p_1, l_1),...,(p_i, l_i),...,(p_n,l_n)\right\}$, one point per type. 
Accordingly, the covariance between $p$ and $l$ in a sample of points is 
$$s_{pl} = \frac{1}{n-1}\left(\sum_{i=1}^n p_i l_i - n \bar{p}\bar{l}\right).$$
Recalling the definition of $L$ (\autoref{eq:mean_type_length}) and noting that $\bar{p} = \frac{1}{n}$ and $\bar{l} = M = L_r$ (recall Property \ref{prop:random_baseline}), we finally obtain
$$s_{pl} = \frac{1}{n-1}(L - L_r).$$

The sample Pearson correlation is  
$$r = \frac{s_{xy}}{s_x s_y},$$
where $s_x$ and $s_y$ are the sample standard deviation of $x$ and $y$, i.e.
\begin{eqnarray*}
s_x = \sqrt{\frac{1}{n-1} \sum_{i=1}^n (x_i - \bar{x})^2} \\
s_y = \sqrt{\frac{1}{n-1} \sum_{i=1}^n (y_i - \bar{y})^2}.
\end{eqnarray*}
Proceeding as we did for the covariance, we find that the Pearson correlation between $p$ and $l$ is 
$$r = \frac{L - L_r}{(n-1)s_p s_l}.$$
Then it is easy to see that $L$ is a linear function of the Pearson correlation $r$ or $s_{pl}$. For instance, 
$$L = ar + b,$$
where 
\begin{eqnarray*}
a = (n-1)s_p s_l \\
b = L_r.
\end{eqnarray*}
Other linear relationships can be shown similarly.

\section{Analysis}
\label{app:analysis}

We here present complementary analyses, tables and plots.

\subsection{The impact of the unsupervised filter}
\label{app:no_filter}

\autoref{tab:coll_comparison_pud} and \autoref{tab:coll_comparison_cv} show the impact of the unsupervised filter in the optional filter. PUD is a controlled setting for the impact of the filter because it is a collection where tokens are of high quality compared to CV. Thus we expect that the impact of the optional filter is low in PUD. Unexpectedly, the number of tokens reduces substantially (a reduction of the order of thousands) in Chinese, Japanese and Korean. An additional drastic reduction in the observed alphabet size in these languages strongly suggests that the optional filter is not adequate for them.  
For these reasons, we believe we should not apply the unsupervised filter to these languages because their writing system is essentially a syllabary. We suspect that the actual need for the exclusion could be a combination of sampling problems relating to a large alphabet size (compared to the Latin script) and a heavy- tailed rank distribution that breaks the optional filter. It is well-known that the rank distribution of Chinese characters is long-tailed, spanning two orders of magnitude \parencite{Deng2014a}, while that of phonemes (the counterpart of letters in many languages using the Latin script) is exponential-like \parencite{Naranan1993,Balasubrahmanyan1996}.
However, that issue should be the subject of future research. 

In CV, we find that the optional filter has a similar impact in languages concerning the reduction in the number of tokens but higher impacts concerning the reduction of the alphabet sizes, suggesting that presence of strings with strange characters. The three languages with the most marked reduction  in alphabet size are French, Spanish, German and Italian, with an alphabet size greater then 100.

\begin{table}[H]
\centering
\caption{The impact of the unsupervised filter in the PUD collection. For every language, we show its linguistic family, the writing system (namely script name according to ISO-15924) and various numeric parameters after applying the mandatory filter but before applying the unsupervised filter, that are $A$, the observed alphabet size (number of distinct characters),
$n$,  the number of types, and, $T$, the number of tokens.
$A'$, $n'$ and $T'$ are the respective values of $A$, $n$ and $T$ after applying the unsupervised filter. 
} 
\label{tab:coll_comparison_pud}
\begin{tabular}{lllrrrrrr}
\hline
Language & Script & Family & $A$ & $A'$ & $n$ & $n'$ & $T$ & $T'$ \\ 
\hline
 Arabic & Arabic & Afro-Asiatic & 47 & 39 & 6600 & 6596 & 18214 & 18201 \\ 
Indonesian & Latin & Austronesian & 39 & 23 & 4596 & 4501 & 16819 & 16702 \\ 
Russian & Cyrillic & Indo-European & 61 & 31 & 7358 & 7113 & 15870 & 15588 \\ 
Hindi & Devanagari & Indo-European & 84 & 50 & 4920 & 4716 & 21184 & 20796 \\ 
Czech & Latin & Indo-European & 49 & 33 & 7360 & 7073 & 15700 & 15331 \\ 
English & Latin & Indo-European & 39 & 25 & 5082 & 5001 & 18135 & 18028 \\ 
French & Latin & Indo-European & 48 & 26 & 5593 & 5214 & 21084 & 20407 \\ 
German & Latin & Indo-European & 39 & 28 & 6215 & 6116 & 18446 & 18331 \\ 
Icelandic & Latin & Indo-European & 43 & 32 & 6175 & 6035 & 16385 & 16209 \\ 
Italian & Latin & Indo-European & 42 & 24 & 5944 & 5606 & 21815 & 21266 \\ 
Polish & Latin & Indo-European & 47 & 31 & 7329 & 7188 & 15386 & 15191 \\ 
Portuguese & Latin & Indo-European & 47 & 38 & 5678 & 5661 & 21873 & 21855 \\ 
Spanish & Latin & Indo-European & 39 & 32 & 5765 & 5750 & 21083 & 21067 \\ 
Swedish & Latin & Indo-European & 39 & 25 & 5842 & 5624 & 16653 & 16378 \\ 
Japanese & Japanese & Japonic & 1549 & 609 & 4990 & 3345 & 24899 & 22538 \\ 
Japanese-strokes & Japanese & Japonic & 1549 & 609 & 4852 & 3345 & 24737 & 22538 \\ 
Japanese-romaji & Latin & Japonic & 23 & 19 & 4984 & 4860 & 24892 & 24743 \\ 
Korean & Hangul & Koreanic & 1002 & 401 & 8031 & 6424 & 14475 & 12540 \\ 
Thai & Thai & Kra-Dai & 89 & 52 & 3818 & 3599 & 21642 & 21121 \\ 
Chinese & Han (Traditional variant) & Sino-Tibetan & 2038 & 814 & 5224 & 3154 & 18129 & 15436 \\ 
Chinese-strokes & Han (Traditional variant) & Sino-Tibetan & 2038 & 814 & 4970 & 3154 & 17845 & 15436 \\ 
Chinese-pinyin & Latin & Sino-Tibetan & 49 & 44 & 5224 & 5038 & 18129 & 17885 \\ 
Turkish & Latin & Turkic & 42 & 28 & 6793 & 6587 & 14092 & 13799 \\ 
Finnish & Latin & Uralic & 39 & 24 & 7076 & 6938 & 12853 & 12701 \\ 
  \hline

\end{tabular}
\end{table}

\begin{table}[H]
\centering
\caption{The impact of the unsupervised filter in the CV collection. The content is the same as in \autoref{tab:coll_comparison_pud}.
'Conlang' stands for 'constructed language', that is an artificially created language. This is not a family in the proper sense as Conlang languages are not related in the common linguistic family sense.
} 
\label{tab:coll_comparison_cv}
\begin{tabular}{lllrrrrrr}
\hline
Language & Script & Family & $A$ & $A'$ & $n$ & $n'$ & $T$ & $T'$ \\ 
\hline
         Arabic & Arabic & Afro-Asiatic & 44 & 31 & 7497 & 6397 & 49448 & 45825 \\ 
        Maltese & Latin & Afro-Asiatic & 40 & 31 & 8148 & 8058 & 44272 & 44112 \\ 
        Vietnamese & Latin & Austroasiatic & 86 & 41 & 574 & 370 & 1300 & 938 \\ 
        Indonesian & Latin & Austronesian & 28 & 22 & 3817 & 3768 & 44336 & 44210 \\ 
        Esperanto & Latin & Conlang & 38 & 27 & 27932 & 27759 & 406725 & 406261 \\ 
        Interlingua & Latin & Conlang & 27 & 20 & 5552 & 5126 & 31428 & 30504 \\ 
        Tamil & Tamil & Dravidian & 44 & 29 & 1525 & 1210 & 7580 & 6439 \\ 
        Persian & Arabic & Indo-European & 105 & 38 & 13240 & 13115 & 1665428 & 1662508 \\ 
        Assamese & Assamese & Indo-European & 60 & 43 & 1115 & 971 & 2000 & 1813 \\ 
        Russian & Cyrillic & Indo-European & 54 & 32 & 31921 & 31827 & 638782 & 637686 \\ 
        Ukrainian & Cyrillic & Indo-European & 44 & 34 & 14399 & 14337 & 120984 & 120760 \\ 
        Panjabi & Devanagari & Indo-European & 48 & 37 & 95 & 84 & 110 & 98 \\ 
        Modern Greek & Greek & Indo-European & 46 & 33 & 5834 & 5813 & 37926 & 37880 \\ 
        Breton & Latin & Indo-European & 41 & 28 & 4322 & 4228 & 38493 & 38237 \\ 
        Catalan & Latin & Indo-European & 67 & 39 & 79213 & 79112 & 3294506 & 3294206 \\ 
        Czech & Latin & Indo-European & 44 & 33 & 16032 & 15518 & 150312 & 147582 \\ 
        Dutch & Latin & Indo-European & 41 & 23 & 10666 & 10225 & 320992 & 316498 \\ 
        English & Latin & Indo-European & 97 & 28 & 173522 & 173023 & 9829660 & 9828713 \\ 
        French & Latin & Indo-European & 244 & 49 & 162740 & 160243 & 3732822 & 3729370 \\ 
        German & Latin & Indo-European & 152 & 30 & 150362 & 148436 & 4235094 & 4230565 \\ 
        Irish & Latin & Indo-European & 31 & 23 & 2311 & 2251 & 22751 & 22593 \\ 
        Italian & Latin & Indo-European & 110 & 34 & 55480 & 54996 & 812604 & 811783 \\ 
        Latvian & Latin & Indo-European & 35 & 27 & 7792 & 7251 & 30358 & 29456 \\ 
        Polish & Latin & Indo-European & 38 & 32 & 25365 & 25340 & 595613 & 595411 \\ 
        Portuguese & Latin & Indo-European & 41 & 27 & 13049 & 11509 & 295042 & 283048 \\ 
        Romanian & Latin & Indo-European & 36 & 29 & 6449 & 6423 & 33370 & 33341 \\ 
        Romansh & Latin & Indo-European & 40 & 26 & 9801 & 9614 & 44192 & 43792 \\ 
        Slovenian & Latin & Indo-European & 28 & 24 & 5994 & 5937 & 26402 & 26304 \\ 
        Spanish & Latin & Indo-European & 186 & 33 & 75617 & 75010 & 1843646 & 1842474 \\ 
        Swedish & Latin & Indo-European & 30 & 25 & 4454 & 4371 & 63282 & 62951 \\ 
        Welsh & Latin & Indo-European & 43 & 22 & 11488 & 11143 & 547345 & 539621 \\ 
        Western Frisian & Latin & Indo-European & 42 & 30 & 8419 & 8383 & 63127 & 63073 \\ 
        Oriya & Odia & Indo-European & 59 & 41 & 921 & 764 & 1929 & 1700 \\ 
        Dhivehi & Thaana & Indo-European & 40 & 27 & 155 & 111 & 1388 & 1284 \\ 
        Georgian & Georgian & Kartvelian & 34 & 25 & 7945 & 6505 & 15481 & 12958 \\ 
        Basque & Latin & Language isolate & 28 & 21 & 24998 & 24748 & 460188 & 458071 \\ 
        Mongolian & Mongolian & Mongolic & 36 & 31 & 14844 & 14608 & 70638 & 70217 \\ 
        Kinyarwanda & Latin & Niger-Congo & 96 & 26 & 135328 & 133815 & 1945038 & 1939810 \\ 
        Abkhazian & Cyrillic & Northwest Caucasian & 37 & 28 & 150 & 119 & 189 & 156 \\ 
        Hakha Chin & Latin & Sino-Tibetan & 28 & 23 & 2515 & 2499 & 17806 & 17776 \\ 
        Chuvash & Cyrillic & Turkic & 36 & 22 & 5565 & 4311 & 16270 & 13583 \\ 
        Kirghiz & Cyrillic & Turkic & 38 & 30 & 10497 & 10130 & 62687 & 61844 \\ 
        Tatar & Cyrillic & Turkic & 47 & 34 & 22313 & 21823 & 145458 & 144356 \\ 
        Yakut & Cyrillic & Turkic & 42 & 28 & 8041 & 7904 & 22795 & 22577 \\ 
        Turkish & Latin & Turkic & 37 & 31 & 8957 & 8926 & 107910 & 107686 \\ 
        Estonian & Latin & Uralic & 34 & 23 & 30135 & 28691 & 123895 & 121549 \\ 
          \hline

\end{tabular}
\end{table}

\subsection{Mean word length and the law of abbreviation}
\label{sec:opt_scores}

In \autoref{tab:opt_scores_pud}, \autoref{tab:opt_scores_cv_characters} and \autoref{tab:opt_scores_cv_meadianDuration}, we show the mean word length ($L$) and the random baseline ($L_r$) as well as the outcome of the correlation test between length and frequency for PUD and for CV when length is measured in characters and also in duration, respectively.

\begin{table}[H]
\centering
\caption{Mean word length and the correlation between frequency and length in PUD. Word length is measured in number of characters. Mean word length ($L$) is followed by the random baseline ($L_r$). Each correlation statistic (Kendall $\tau$ or Pearson $r$) is followed by \textit{p}-values after applying Holm-Bonferroni correction (rather than being the direct output of the correlation test).
}  
\label{tab:opt_scores_pud}
\begin{tabular}{lllrrrlrl}
  \hline
language & family & script & $L$ & $L_r$ & $\tau$ & $\tau_{pvalue}$ & $r$ & $r_{pvalue}$\\  
  \hline
 Arabic & Afro-Asiatic & Arabic & 4.03 & 5.54 & -0.13 & $8.32 \times 10^{-32}$ & -0.13 & $1.12 \times 10^{-20}$ \\ 
  Czech & Indo-European & Latin & 5.44 & 7.27 & -0.22 & $1.20 \times 10^{-113}$ & -0.15 & $2.47 \times 10^{-36}$ \\ 
  English & Indo-European & Latin & 4.87 & 7.00 & -0.20 & $2.52 \times 10^{-66}$ & -0.12 & $6.98 \times 10^{-17}$ \\ 
  French & Indo-European & Latin & 4.81 & 7.47 & -0.16 & $2.44 \times 10^{-49}$ & -0.12 & $4.24 \times 10^{-19}$ \\ 
  German & Indo-European & Latin & 5.74 & 8.56 & -0.23 & $1.25 \times 10^{-108}$ & -0.12 & $3.85 \times 10^{-21}$ \\ 
  Indonesian & Austronesian & Latin & 5.96 & 7.35 & -0.11 & $6.37 \times 10^{-21}$ & -0.12 & $6.53 \times 10^{-15}$ \\ 
  Italian & Indo-European & Latin & 4.85 & 7.64 & -0.16 & $4.09 \times 10^{-54}$ & -0.13 & $8.45 \times 10^{-23}$ \\ 
  Polish & Indo-European & Latin & 6.07 & 8.00 & -0.19 & $1.12 \times 10^{-80}$ & -0.13 & $2.78 \times 10^{-26}$ \\ 
  Portuguese & Indo-European & Latin & 4.35 & 7.47 & -0.20 & $9.96 \times 10^{-67}$ & -0.12 & $1.12 \times 10^{-17}$ \\ 
  Russian & Indo-European & Cyrillic & 6.04 & 8.08 & -0.19 & $4.58 \times 10^{-88}$ & -0.13 & $4.85 \times 10^{-26}$ \\ 
  Spanish & Indo-European & Latin & 4.83 & 7.59 & -0.16 & $4.10 \times 10^{-51}$ & -0.11 & $1.89 \times 10^{-17}$ \\ 
  Swedish & Indo-European & Latin & 5.41 & 7.99 & -0.23 & $3.99 \times 10^{-101}$ & -0.13 & $6.28 \times 10^{-21}$ \\ 
  Turkish & Turkic & Latin & 6.43 & 7.94 & -0.24 & $4.26 \times 10^{-124}$ & -0.12 & $4.20 \times 10^{-23}$ \\ 
   \hline

\end{tabular}
\end{table}

\begin{table}[H]
\centering
\caption{Mean word length and the correlation between frequency and length in CV. Word length is measured in number of characters. Content is the same as in \ref{tab:opt_scores_pud}. 
'Conlang' stands for 'constructed language', that is an artificially created language. This is not a family in the proper sense, and Conlang languages are not related in the common family sense. 
}
\label{tab:opt_scores_cv_characters}
\begin{tabular}{lllrrrlrl}
  \hline
language & family & script & $L$ & $L_r$ & $\tau$ & $\tau_{pvalue}$ & $r$ & $r_{pvalue}$\\ 
  \hline
 Abkhazian & Northwest Caucasian & Cyrillic & 5.94 & 6.42 & -0.32 & $4.48 \times 10^{-5}$ & -0.29 & $1.43 \times 10^{-3}$ \\ 
  Arabic & Afro-Asiatic & Arabic & 4.10 & 5.06 & -0.14 & $5.32 \times 10^{-43}$ & -0.14 & $2.04 \times 10^{-28}$ \\ 
  Assamese & Indo-European & Assamese & 4.57 & 5.36 & -0.31 & $4.73 \times 10^{-31}$ & -0.27 & $3.09 \times 10^{-17}$ \\ 
  Basque & Language isolate & Latin & 6.41 & 8.89 & -0.16 & $2.68 \times 10^{-262}$ & -0.11 & $6.95 \times 10^{-69}$ \\ 
  Breton & Indo-European & Latin & 3.97 & 6.31 & -0.24 & $4.93 \times 10^{-86}$ & -0.19 & $4.09 \times 10^{-35}$ \\ 
  Catalan & Indo-European & Latin & 4.90 & 8.58 & -0.15 & $0.00$ & -0.05 & $9.53 \times 10^{-51}$ \\ 
  Chuvash & Turkic & Cyrillic & 6.00 & 7.35 & -0.22 & $5.49 \times 10^{-74}$ & -0.21 & $3.80 \times 10^{-43}$ \\ 
  Czech & Indo-European & Latin & 4.83 & 7.17 & -0.22 & $1.75 \times 10^{-295}$ & -0.13 & $1.69 \times 10^{-58}$ \\ 
  Dhivehi & Indo-European & Thaana & 3.32 & 7.61 & -0.16 & $1.65 \times 10^{-2}$ & -0.31 & $1.24 \times 10^{-3}$ \\ 
  Dutch & Indo-European & Latin & 4.72 & 8.26 & -0.28 & $0.00$ & -0.10 & $1.35 \times 10^{-24}$ \\ 
  English & Indo-European & Latin & 4.61 & 7.79 & -0.07 & $0.00$ & -0.03 & $3.45 \times 10^{-45}$ \\ 
  Esperanto & Conlang & Latin & 4.83 & 7.73 & -0.18 & $0.00$ & -0.08 & $1.12 \times 10^{-41}$ \\ 
  Estonian & Uralic & Latin & 6.16 & 8.85 & -0.24 & $0.00$ & -0.09 & $2.55 \times 10^{-48}$ \\ 
  French & Indo-European & Latin & 5.04 & 8.13 & -0.04 & $3.57 \times 10^{-85}$ & -0.04 & $8.56 \times 10^{-46}$ \\ 
  Georgian & Kartvelian & Georgian & 7.17 & 8.22 & -0.12 & $3.67 \times 10^{-31}$ & -0.10 & $3.47 \times 10^{-15}$ \\ 
  German & Indo-European & Latin & 5.73 & 10.30 & -0.12 & $0.00$ & -0.04 & $4.21 \times 10^{-59}$ \\ 
  Hakha Chin & Sino-Tibetan & Latin & 3.29 & 5.29 & -0.29 & $4.31 \times 10^{-72}$ & -0.15 & $3.88 \times 10^{-13}$ \\ 
  Indonesian & Austronesian & Latin & 5.37 & 7.24 & -0.20 & $1.13 \times 10^{-59}$ & -0.16 & $2.73 \times 10^{-21}$ \\ 
  Interlingua & Conlang & Latin & 4.43 & 7.43 & -0.24 & $8.95 \times 10^{-101}$ & -0.16 & $7.39 \times 10^{-31}$ \\ 
  Irish & Indo-European & Latin & 4.20 & 6.58 & -0.21 & $2.38 \times 10^{-41}$ & -0.17 & $5.18 \times 10^{-15}$ \\ 
  Italian & Indo-European & Latin & 5.29 & 8.16 & -0.06 & $2.24 \times 10^{-67}$ & -0.06 & $4.19 \times 10^{-49}$ \\ 
  Kinyarwanda & Niger-Congo & Latin & 6.13 & 9.20 & -0.19 & $0.00$ & -0.06 & $3.32 \times 10^{-117}$ \\ 
  Kirghiz & Turkic & Cyrillic & 6.01 & 7.78 & -0.19 & $1.45 \times 10^{-141}$ & -0.16 & $6.13 \times 10^{-57}$ \\ 
  Latvian & Indo-European & Latin & 4.79 & 7.09 & -0.26 & $5.81 \times 10^{-160}$ & -0.18 & $1.36 \times 10^{-53}$ \\ 
  Maltese & Afro-Asiatic & Latin & 5.07 & 7.35 & -0.20 & $2.32 \times 10^{-107}$ & -0.14 & $1.58 \times 10^{-36}$ \\ 
  Modern Greek & Indo-European & Greek & 4.85 & 7.64 & -0.24 & $3.73 \times 10^{-124}$ & -0.16 & $1.77 \times 10^{-34}$ \\ 
  Mongolian & Mongolic & Mongolian & 5.47 & 7.31 & -0.23 & $1.73 \times 10^{-263}$ & -0.15 & $2.31 \times 10^{-76}$ \\ 
  Oriya & Indo-European & Odia & 4.21 & 5.35 & -0.33 & $2.00 \times 10^{-28}$ & -0.31 & $2.94 \times 10^{-17}$ \\ 
  Panjabi & Indo-European & Devanagari & 3.68 & 3.88 & -0.32 & $8.69 \times 10^{-4}$ & -0.26 & $8.60 \times 10^{-3}$ \\ 
  Persian & Indo-European & Arabic & 3.80 & 5.49 & -0.21 & $2.38 \times 10^{-229}$ & -0.12 & $2.06 \times 10^{-45}$ \\ 
  Polish & Indo-European & Latin & 5.27 & 7.87 & -0.17 & $8.03 \times 10^{-292}$ & -0.10 & $2.47 \times 10^{-58}$ \\ 
  Portuguese & Indo-European & Latin & 4.53 & 7.49 & -0.19 & $1.09 \times 10^{-168}$ & -0.13 & $1.21 \times 10^{-41}$ \\ 
  Romanian & Indo-European & Latin & 5.03 & 7.67 & -0.21 & $3.27 \times 10^{-97}$ & -0.17 & $6.46 \times 10^{-41}$ \\ 
  Romansh & Indo-European & Latin & 4.94 & 7.56 & -0.24 & $5.91 \times 10^{-184}$ & -0.15 & $5.42 \times 10^{-48}$ \\ 
  Russian & Indo-European & Cyrillic & 6.31 & 9.00 & -0.13 & $7.75 \times 10^{-225}$ & -0.09 & $3.03 \times 10^{-52}$ \\ 
  Slovenian & Indo-European & Latin & 4.56 & 6.43 & -0.21 & $1.47 \times 10^{-88}$ & -0.15 & $4.71 \times 10^{-29}$ \\ 
  Spanish & Indo-European & Latin & 5.01 & 7.92 & -0.03 & $5.95 \times 10^{-32}$ & -0.04 & $3.48 \times 10^{-29}$ \\ 
  Swedish & Indo-European & Latin & 4.04 & 6.87 & -0.28 & $6.91 \times 10^{-129}$ & -0.15 & $1.62 \times 10^{-22}$ \\ 
  Tamil & Dravidian & Tamil & 5.68 & 7.08 & -0.28 & $1.01 \times 10^{-35}$ & -0.23 & $5.21 \times 10^{-16}$ \\ 
  Tatar & Turkic & Cyrillic & 5.41 & 7.45 & -0.24 & $0.00$ & -0.16 & $3.15 \times 10^{-118}$ \\ 
  Turkish & Turkic & Latin & 6.00 & 8.09 & -0.22 & $1.32 \times 10^{-158}$ & -0.16 & $2.51 \times 10^{-48}$ \\ 
  Ukrainian & Indo-European & Cyrillic & 5.52 & 7.67 & -0.16 & $3.01 \times 10^{-136}$ & -0.14 & $1.74 \times 10^{-61}$ \\ 
  Vietnamese & Austroasiatic & Latin & 3.24 & 3.47 & -0.19 & $2.98 \times 10^{-5}$ & -0.20 & $1.96 \times 10^{-4}$ \\ 
  Welsh & Indo-European & Latin & 4.17 & 7.05 & -0.21 & $2.40 \times 10^{-185}$ & -0.12 & $4.39 \times 10^{-38}$ \\ 
  Western Frisian & Indo-European & Latin & 4.38 & 7.99 & -0.29 & $1.19 \times 10^{-244}$ & -0.13 & $2.62 \times 10^{-33}$ \\ 
  Yakut & Turkic & Cyrillic & 6.32 & 7.99 & -0.26 & $5.48 \times 10^{-185}$ & -0.19 & $2.12 \times 10^{-65}$ \\ 
   \hline

\end{tabular}
\end{table}

\begin{table}[H]
\centering
\caption{Mean word length and the correlation between frequency and length in CV. Word length is measured in duration. Content is the same as in \ref{tab:opt_scores_cv_characters}.} 
\label{tab:opt_scores_cv_meadianDuration}
\begin{tabular}{lllrrrlrl}
  \hline
language & family & script & $L$ & $L_r$ & $\tau$ & $\tau_{pvalue}$ & $r$ & $r_{pvalue}$\\ 
  \hline
 Abkhazian & Northwest Caucasian & Cyrillic & 0.74 & 0.81 & -0.20 & $1.23 \times 10^{-2}$ & -0.21 & $2.52 \times 10^{-2}$ \\ 
  Arabic & Afro-Asiatic & Arabic & 0.46 & 0.58 & -0.12 & $1.75 \times 10^{-40}$ & -0.15 & $2.00 \times 10^{-31}$ \\ 
  Assamese & Indo-European & Assamese & 0.43 & 0.50 & -0.22 & $1.25 \times 10^{-17}$ & -0.19 & $3.14 \times 10^{-9}$ \\ 
  Basque & Language isolate & Latin & 0.44 & 0.63 & -0.21 & $0.00$ & -0.12 & $1.29 \times 10^{-78}$ \\ 
  Breton & Indo-European & Latin & 0.31 & 0.51 & -0.25 & $1.92 \times 10^{-107}$ & -0.20 & $4.94 \times 10^{-39}$ \\ 
  Catalan & Indo-European & Latin & 0.35 & 0.68 & -0.21 & $0.00$ & -0.06 & $8.70 \times 10^{-69}$ \\ 
  Chuvash & Turkic & Cyrillic & 0.44 & 0.54 & -0.26 & $1.18 \times 10^{-116}$ & -0.22 & $6.89 \times 10^{-49}$ \\ 
  Czech & Indo-European & Latin & 0.37 & 0.57 & -0.21 & $6.40 \times 10^{-295}$ & -0.14 & $5.07 \times 10^{-70}$ \\ 
  Dhivehi & Indo-European & Thaana & 0.32 & 0.71 & -0.17 & $2.40 \times 10^{-2}$ & -0.24 & $1.51 \times 10^{-2}$ \\ 
  Dutch & Indo-European & Latin & 0.29 & 0.55 & -0.28 & $0.00$ & -0.12 & $1.47 \times 10^{-33}$ \\ 
  English & Indo-European & Latin & 0.33 & 0.67 & -0.17 & $0.00$ & -0.04 & $4.83 \times 10^{-62}$ \\ 
  Esperanto & Conlang & Latin & 0.49 & 0.81 & -0.18 & $0.00$ & -0.09 & $1.25 \times 10^{-47}$ \\ 
  Estonian & Uralic & Latin & 0.39 & 0.58 & -0.23 & $0.00$ & -0.09 & $4.65 \times 10^{-55}$ \\ 
  French & Indo-European & Latin & 0.32 & 0.63 & -0.21 & $0.00$ & -0.04 & $7.25 \times 10^{-71}$ \\ 
  Georgian & Kartvelian & Georgian & 0.52 & 0.61 & -0.15 & $6.51 \times 10^{-51}$ & -0.12 & $9.17 \times 10^{-21}$ \\ 
  German & Indo-European & Latin & 0.37 & 0.76 & -0.22 & $0.00$ & -0.05 & $1.57 \times 10^{-96}$ \\ 
  Hakha Chin & Sino-Tibetan & Latin & 0.29 & 0.44 & -0.25 & $9.56 \times 10^{-64}$ & -0.14 & $1.10 \times 10^{-11}$ \\ 
  Indonesian & Austronesian & Latin & 0.38 & 0.52 & -0.22 & $1.29 \times 10^{-76}$ & -0.17 & $8.41 \times 10^{-26}$ \\ 
  Interlingua & Conlang & Latin & 0.40 & 0.69 & -0.24 & $9.77 \times 10^{-114}$ & -0.18 & $3.80 \times 10^{-36}$ \\ 
  Irish & Indo-European & Latin & 0.30 & 0.47 & -0.24 & $1.42 \times 10^{-55}$ & -0.19 & $1.02 \times 10^{-19}$ \\ 
  Italian & Indo-European & Latin & 0.38 & 0.65 & -0.19 & $0.00$ & -0.08 & $4.19 \times 10^{-87}$ \\ 
  Kinyarwanda & Niger-Congo & Latin & 0.44 & 0.72 & -0.21 & $0.00$ & -0.06 & $7.54 \times 10^{-101}$ \\ 
  Kirghiz & Turkic & Cyrillic & 0.44 & 0.57 & -0.20 & $1.38 \times 10^{-159}$ & -0.16 & $1.63 \times 10^{-55}$ \\ 
  Latvian & Indo-European & Latin & 0.39 & 0.59 & -0.23 & $1.70 \times 10^{-141}$ & -0.19 & $4.58 \times 10^{-60}$ \\ 
  Maltese & Afro-Asiatic & Latin & 0.35 & 0.54 & -0.21 & $9.96 \times 10^{-140}$ & -0.15 & $3.89 \times 10^{-42}$ \\ 
  Modern Greek & Indo-European & Greek & 0.38 & 0.63 & -0.21 & $3.20 \times 10^{-105}$ & -0.17 & $1.46 \times 10^{-37}$ \\ 
  Mongolian & Mongolic & Mongolian & 0.36 & 0.48 & -0.25 & $0.00$ & -0.15 & $3.12 \times 10^{-73}$ \\ 
  Oriya & Indo-European & Odia & 0.39 & 0.49 & -0.33 & $2.21 \times 10^{-31}$ & -0.32 & $1.59 \times 10^{-18}$ \\ 
  Panjabi & Indo-European & Devanagari & 0.70 & 0.73 & -0.18 & $4.63 \times 10^{-2}$ & -0.15 & $8.44 \times 10^{-2}$ \\ 
  Persian & Indo-European & Arabic & 0.36 & 0.54 & -0.25 & $0.00$ & -0.14 & $4.66 \times 10^{-58}$ \\ 
  Polish & Indo-European & Latin & 0.38 & 0.57 & -0.17 & $0.00$ & -0.12 & $2.88 \times 10^{-82}$ \\ 
  Portuguese & Indo-European & Latin & 0.35 & 0.61 & -0.22 & $1.15 \times 10^{-243}$ & -0.15 & $4.38 \times 10^{-59}$ \\ 
  Romanian & Indo-European & Latin & 0.36 & 0.57 & -0.23 & $2.49 \times 10^{-127}$ & -0.17 & $2.82 \times 10^{-43}$ \\ 
  Romansh & Indo-European & Latin & 0.41 & 0.66 & -0.26 & $7.70 \times 10^{-248}$ & -0.17 & $2.06 \times 10^{-64}$ \\ 
  Russian & Indo-European & Cyrillic & 0.42 & 0.60 & -0.15 & $2.13 \times 10^{-299}$ & -0.10 & $5.30 \times 10^{-75}$ \\ 
  Slovenian & Indo-European & Latin & 0.44 & 0.63 & -0.25 & $3.37 \times 10^{-146}$ & -0.17 & $3.04 \times 10^{-40}$ \\ 
  Spanish & Indo-European & Latin & 0.36 & 0.62 & -0.14 & $0.00$ & -0.05 & $2.05 \times 10^{-41}$ \\ 
  Swedish & Indo-European & Latin & 0.27 & 0.52 & -0.29 & $1.03 \times 10^{-156}$ & -0.18 & $4.76 \times 10^{-32}$ \\ 
  Tamil & Dravidian & Tamil & 0.54 & 0.66 & -0.31 & $2.06 \times 10^{-48}$ & -0.22 & $5.35 \times 10^{-14}$ \\ 
  Tatar & Turkic & Cyrillic & 0.38 & 0.52 & -0.26 & $0.00$ & -0.17 & $8.68 \times 10^{-141}$ \\ 
  Turkish & Turkic & Latin & 0.41 & 0.54 & -0.21 & $7.11 \times 10^{-158}$ & -0.16 & $9.43 \times 10^{-50}$ \\ 
  Ukrainian & Indo-European & Cyrillic & 0.43 & 0.59 & -0.18 & $3.01 \times 10^{-176}$ & -0.16 & $3.53 \times 10^{-86}$ \\ 
  Vietnamese & Austroasiatic & Latin & 0.29 & 0.33 & -0.07 & $4.63 \times 10^{-2}$ & -0.14 & $1.40 \times 10^{-2}$ \\ 
  Welsh & Indo-European & Latin & 0.32 & 0.58 & -0.20 & $6.25 \times 10^{-197}$ & -0.15 & $3.21 \times 10^{-58}$ \\ 
  Western Frisian & Indo-European & Latin & 0.32 & 0.61 & -0.31 & $0.00$ & -0.15 & $8.59 \times 10^{-43}$ \\ 
  Yakut & Turkic & Cyrillic & 0.43 & 0.54 & -0.25 & $2.41 \times 10^{-186}$ & -0.18 & $9.50 \times 10^{-58}$ \\ 
   \hline

\end{tabular}
\end{table}



\end{appendices}

\end{document}